\documentclass[letterpaper, 10 pt, conference]{ieeeconf}
\IEEEoverridecommandlockouts 
\let\labelindent\relax
\usepackage{hyperref, url, amsmath, amssymb, booktabs, graphicx, bm, color, algorithm, tabularx, enumitem}
\usepackage[noend]{algpseudocode}
\graphicspath{{images/}}

\newcommand{\new}[1]{#1}

\newcolumntype{Y}{>{\centering\arraybackslash}X}

\title{\LARGE \bf Assistive VR Gym: Interactions with Real People \\ to Improve Virtual Assistive Robots}

\author{Zackory Erickson$^{*}$, Yijun Gu$^{*}$, and Charles C. Kemp
\thanks{Zackory Erickson, Yijun Gu, and Charles C. Kemp are with the Healthcare Robotics Lab, Georgia Institute of Technology, Atlanta, GA., USA.}%
\thanks{Zackory Erickson is the corresponding author {\tt\footnotesize zackory@gatech.edu}.}%
\thanks{$^{*}$ Equal contribution.}%
}

\begin{document}

\maketitle
\thispagestyle{empty}
\pagestyle{empty}
\begin{abstract}
Versatile robotic caregivers could benefit millions of people worldwide, including older adults and people with disabilities. Recent work has explored how robotic caregivers can learn to interact with people through physics simulations, yet transferring what has been learned to real robots remains challenging. Virtual reality (VR) has the potential to help bridge the gap between simulations and the real world. We present Assistive VR Gym (AVR Gym), which enables real people to interact with virtual assistive robots. We also provide evidence that AVR Gym can help researchers improve the performance of simulation-trained assistive robots with real people. Prior to AVR Gym, we trained robot control policies (\emph{Original Policies}) solely in simulation for four robotic caregiving tasks (robot-assisted feeding, drinking, itch scratching, and bed bathing) with two simulated robots (PR2 from Willow Garage and Jaco from Kinova). With AVR Gym, we developed \emph{Revised Policies} based on insights gained from testing the Original policies with real people. Through a formal study with eight participants in AVR Gym, we found that the Original policies performed poorly, the Revised policies performed significantly better, and that improvements to the biomechanical models used to train the Revised policies resulted in simulated people that better match real participants. Notably, participants significantly disagreed that the Original policies were successful at assistance, but significantly agreed that the Revised policies were successful at assistance. Overall, our results suggest that VR can be used to improve the performance of simulation-trained control policies with real people without putting people at risk, thereby serving as a valuable stepping stone to real robotic assistance. 
\end{abstract}

\section{Introduction}
\label{sec:intro}

Robotic assistance with activities of daily living (ADLs) could increase the independence of people with disabilities, improve quality of life, and help address pressing societal needs, such as aging populations, high healthcare costs, and shortages of healthcare workers~\cite{taylor2018americans, section2011retooling}. 

Physics simulations provide an opportunity for robots to safely learn how to physically assist people. Yet, the reality gap between physics simulations and real assistance poses several challenges. By bringing real people into the robot's virtual world, virtual reality (VR) has the potential to serve as a stepping stone between simulated worlds and the real world. For assistive robots, the person receiving assistance is an extremely important and complex part of the environment. As we show through a formal study, enabling real people to interact with a virtual robot can quickly reveal deficiencies through both objective and subjective measures of performance. 

\begin{figure}
\centering
\includegraphics[width=0.23\textwidth, trim={20cm 0cm 12cm 0cm}, clip]{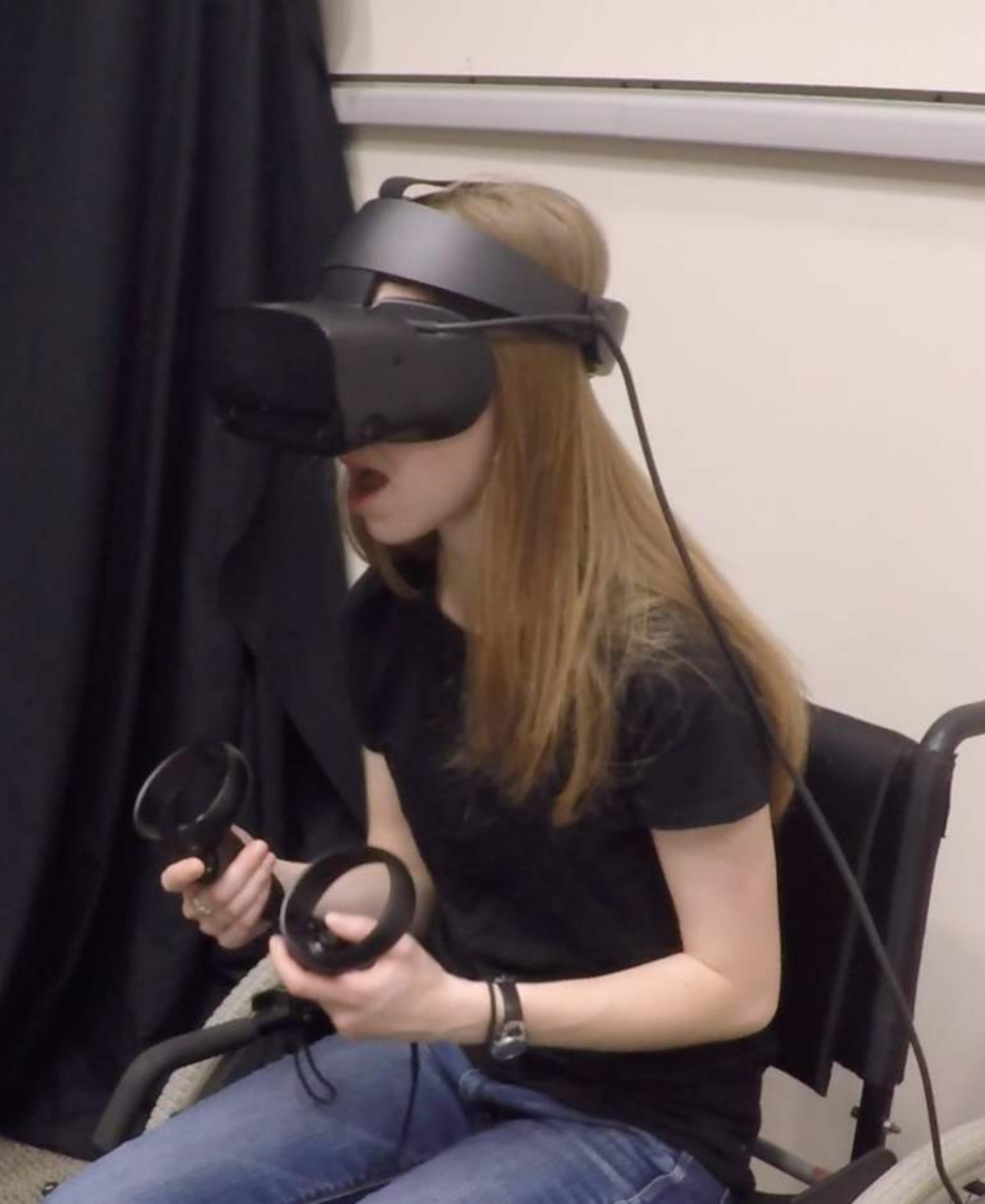}
\includegraphics[width=0.23\textwidth, trim={20cm 0cm 12cm 0cm}, clip]{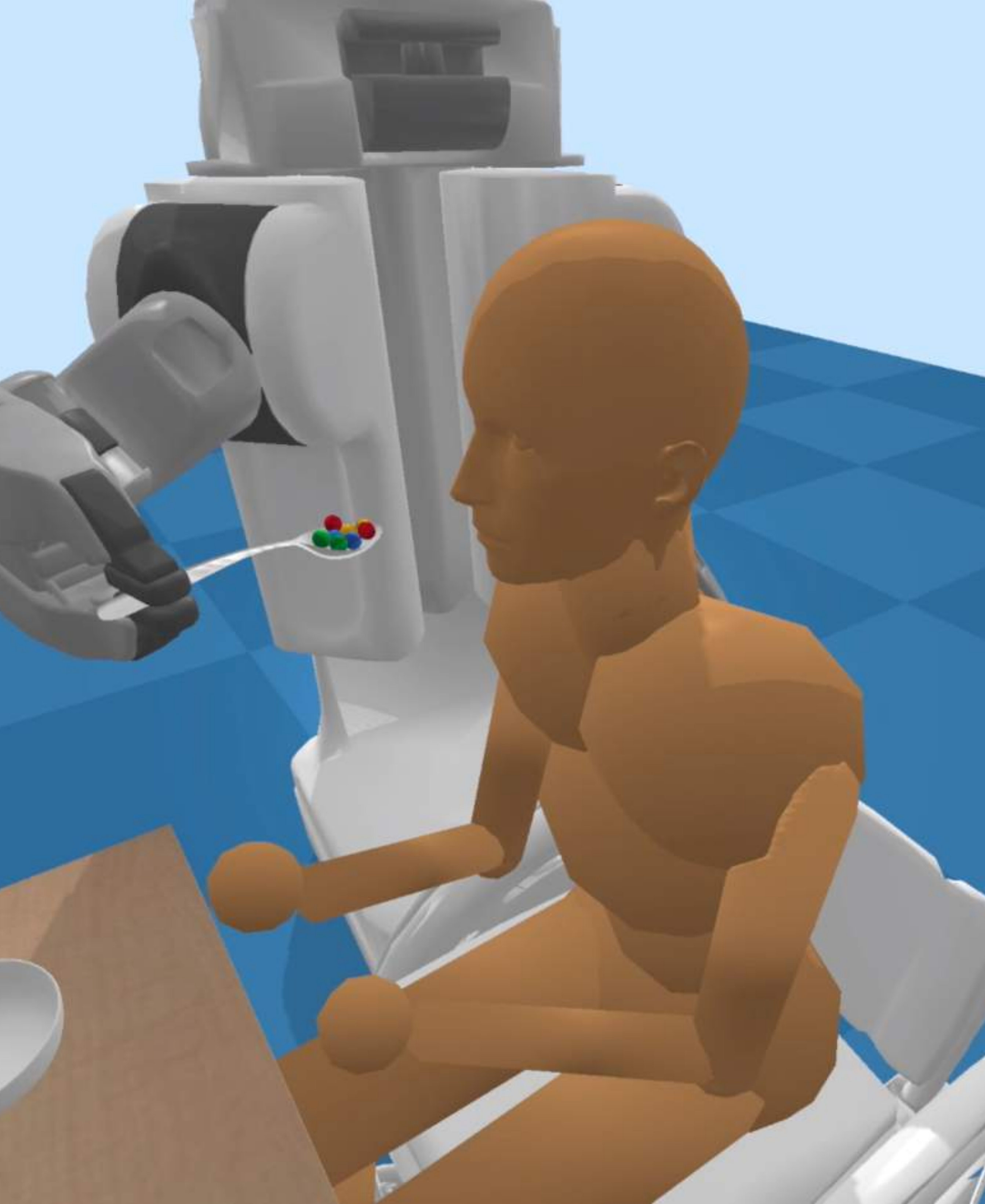}
\caption{\label{fig:intro}A participant using virtual reality to interact with a PR2 robot that has learned how to provide feeding assistance through physics simulation.}
\vspace{-0.4cm}
\end{figure}

To facilitate the use of VR in the development of assistive robots, we present Assistive VR Gym\footnote{\scriptsize\url{https://github.com/Healthcare-Robotics/assistive-vr-gym}} (AVR Gym) an open source framework that enables real people to interact with virtual assistive robots within a physics simulation (see Fig.~\ref{fig:intro}). AVR Gym builds on Assistive Gym, which is an open source physics simulation framework for assistive robots that models multiple assistive tasks~\cite{erickson2019assistive}. AVR Gym enables people to interact with virtual robots without putting themselves at risk, which is especially valuable when evaluating controllers that have been trained in simulation with virtual humans. 

As confirmed through a formal study with eight participants, our use of AVR Gym enabled us to identify significant, unexpected shortcomings in the simulation-trained baseline control policies that were originally released with Assistive Gym. Moreover, we were able to use AVR Gym to improve the control policies by discovering that the simulated humans did not adequately represent the biomechanics of real people. Improving the simulated humans used to train control policies, resulted in Revised control policies that dramatically outperformed our Original policies. 

In this paper, we make the following contributions:
\begin{itemize}
\item Present Assistive VR Gym (AVR Gym), an open source framework that enables real people to interact with virtual assistive robots.
\item Present experimental methods, including objective and subjective measures, to evaluate virtual assistive robots.
\item Provide evidence that VR can be used to improve performance of simulation-trained robot control policies.
\item Provide evidence that biomechanical models significantly impact policy performance and can be assessed and improved via VR.
\end{itemize}

\section{Related Work}
\label{sec:related_work}
\subsection{Physically Assistive Robotics}

A significant amount of research has been conducted on real robotic systems for providing physical assistance to people. For example, researchers have demonstrated how table-mounted robotic arms can provide drinking assistance to people using a force sensing smart cup~\cite{goldau2019autonomous} and a brain-machine interface for shared-autonomy~\cite{schroer2015autonomous}. Similarly, robots have shown promising results for providing feeding assistance~\cite{park2016towards, canal2016personalization, perera2017eeg, park2018multimodal, rhodes2018robot, bhattacharjee2020more}.
Prior research has also investigated robotic assistance for bed bathing~\cite{king2010towards, erickson2019multidimensional}, which can be especially valuable for people who are confined to a bed due to disabilities or injuries. Robots also present an opportunity to help individuals in getting dressed with a variety of garments. Robot-assisted dressing has been demonstrated on several robotic platforms for assisting both able-bodied participants and real people with disabilities~\cite{klee2015personalized, joshi2017robotic, canal2018joining, jevtic2018personalized, erickson2018tracking, erickson2019multidimensional, chance2018elbows, pignat2017learning, koganti2016bayesian, gao2016iterative}.

Despite these promising results in physical robotic assistance, it remains challenging to design robotic systems that can assist with multiple tasks across a wide spectrum of human shapes, sizes, weights, and disabilities. This is due in part to difficulties and costs associated with evaluating assistive robotic systems across a large distribution of people. 
Physics simulation presents an opportunity for robots to learn to safely interact with people over many tasks and environments. For instance, researchers have demonstrated the use of physics simulation for learning robot controllers for robot-assisted dressing tasks~\cite{erickson2017does, yu2017haptic, erickson2018deep, kapusta2019personalized}. Clegg et al. has also used Assistive Gym to learn control policies for a real PR2 robot to dress an arm of a humanoid robot with a hospital gown~\cite{clegg2020learning}.

\subsection{Virtual Reality}
We use virtual reality to evaluate and improve simulation-trained assistive robots with real people.
Studies have provided evidence that virtual reality can offer people both a sense of presence and embodiment~\cite{banos2004immersion, kilteni2012sense}.
Accordingly, VR provides an opportunity to assess important attributes for human-robot interaction (HRI), including proxemics, legibility of motion, and embodiment~\cite{takayama2009influences, wainer2006role}.

Virtual reality has been widely used as a tool to improve robot performance across an assortment of tasks. A common use for virtual reality is for teleoperation of robots for dexterous manipulation tasks~\cite{kumar2015mujoco, lipton2017baxter}.
This virtual reality teleoperation has also been used to collect high-quality robotic manipulation demonstrations~\cite{kumar2016learning, zhang2018deep, liu2019high}. 
Within robotic rehabilitation, virtual reality has been used to enhance rehabilitation training and assess rehabilitation metrics in real time~\cite{frisoli2009robotic, brutsch2011virtual, ballester2015accelerating, huang2018combined, bernardoni2019virtual}.

Across these works, virtual reality has often been used for people to take control of robots; to provide demonstrations and to train robots for performing various tasks. In this paper, we show that virtual reality can be used to safely evaluate and improve simulation-trained assistive robot controllers with real people.

\begin{figure*}
\centering
\includegraphics[width=0.32\textwidth, trim={2cm 0.5cm 0cm 2.5cm}, clip]{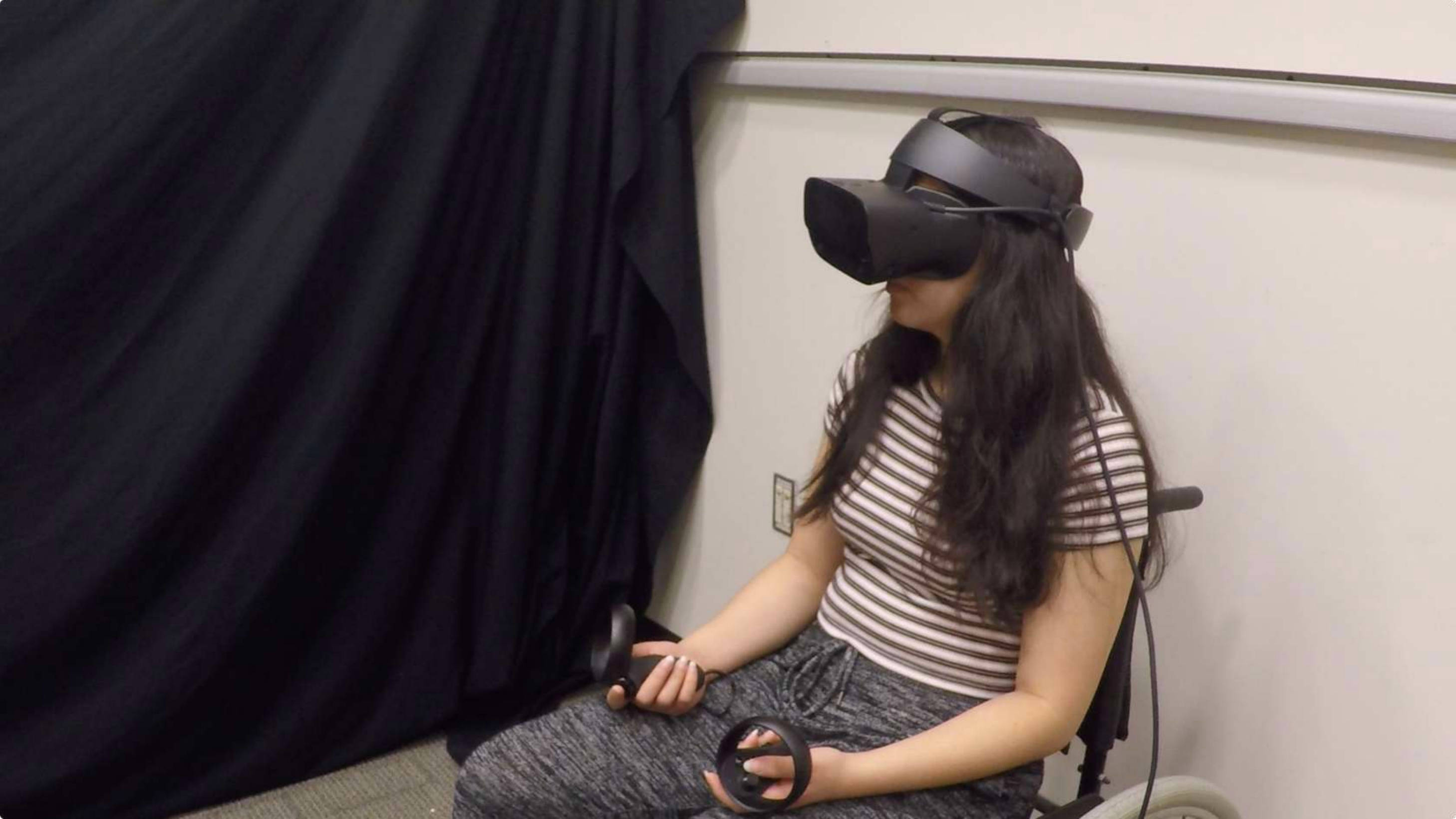}
\includegraphics[width=0.32\textwidth, trim={2cm 0.5cm 0cm 2.5cm}, clip]{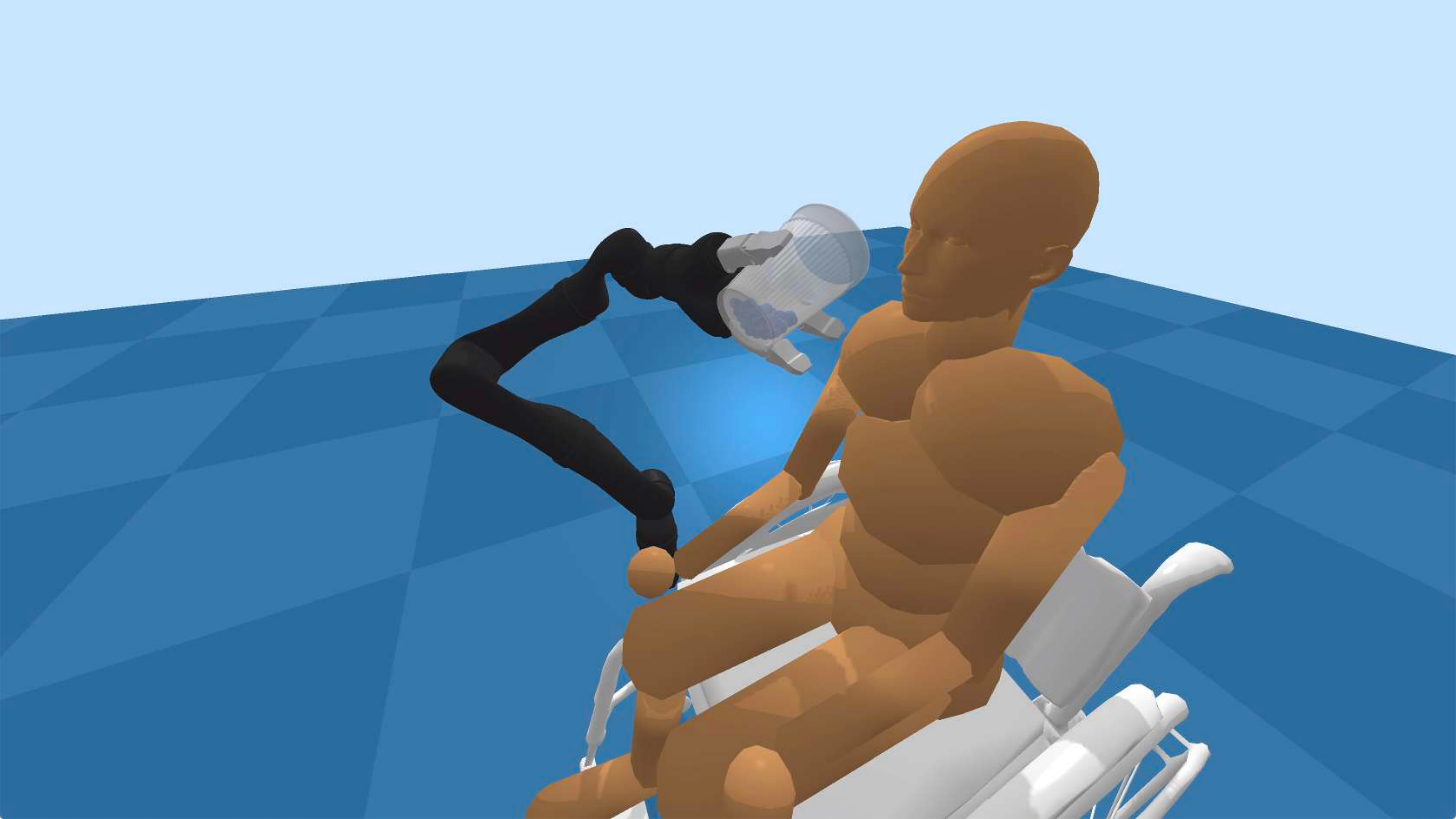}
\includegraphics[width=0.32\textwidth, trim={1cm 0.5cm 1cm 2.5cm}, clip]{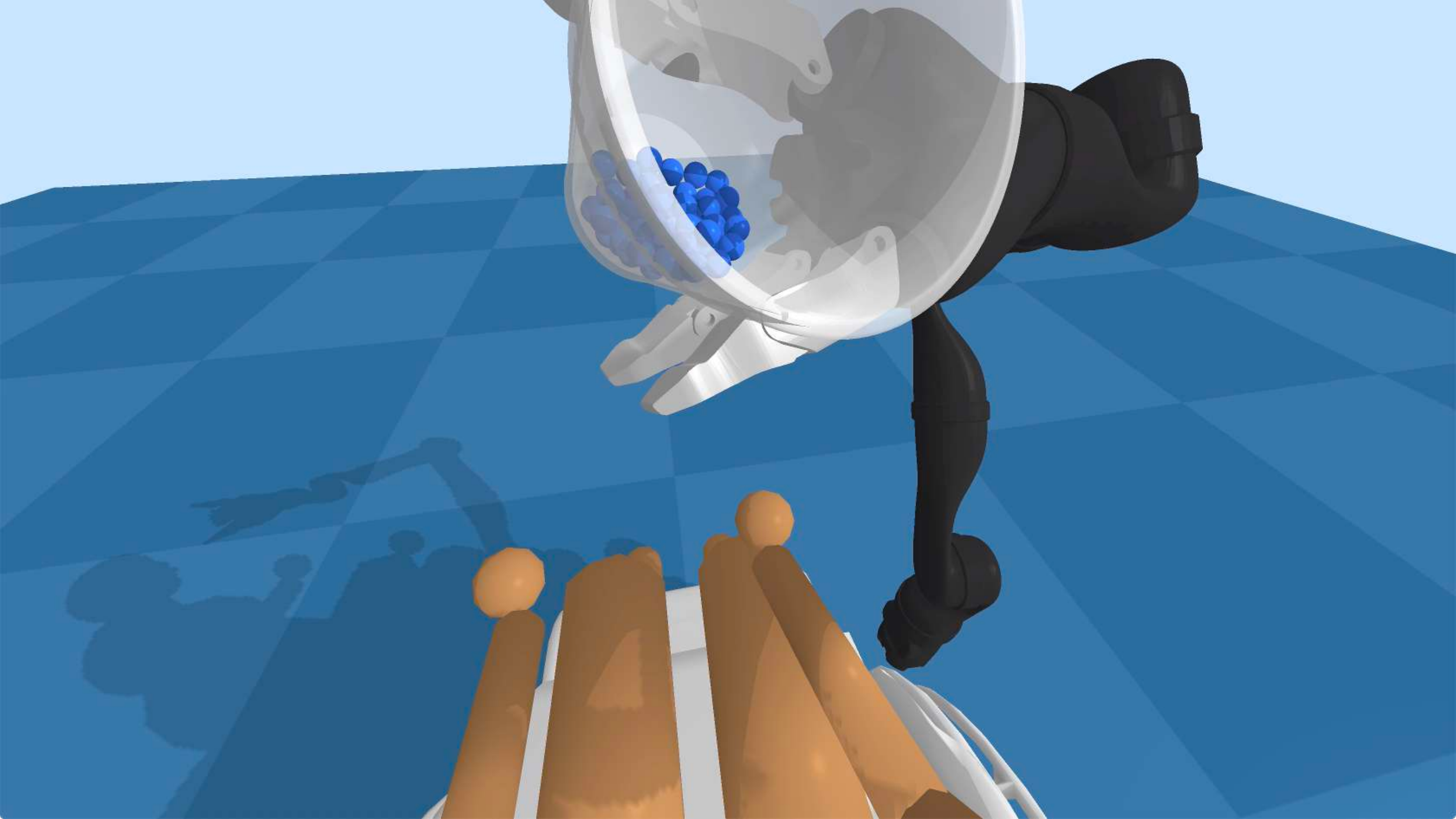}
\vspace{-0.2cm}
\caption{\label{fig:vr}(Left) Human participant using the virtual reality interface to receive drinking assistance from a simulated robot. (Middle) 3rd person perspective of the simulated human model that the human participant is controlling. (Right) 1st person perspective of what a participant observed in virtual reality.}
\vspace{-0.3cm}
\end{figure*}

\section{Assistive VR Gym}
\label{sec:vr}

AVR Gym builds on Assistive Gym and is implemented using the open source PyBullet physics engine~\cite{coumanspybullet}. We connect assistive environments from Assistive Gym into virtual reality with vrBullet, a virtual reality physics server that uses the OpenVR API. We use the Oculus Rift S virtual reality head-mounted display, which uses inside-out tracking, and two Oculus Touch controllers, each of which provide 3-DoF position and 3-DoF orientation tracking. Fig.~\ref{fig:vr} depicts the virtual reality setup with 3rd and 1st person perspectives of the virtual environment.

The simulated human model in virtual reality has 20 controllable joints including two 7-DoF arms, a 3-DoF head, and a 3-DoF waist. Using the Oculus' inside-out tracking, we can estimate the height of each human participant as the difference from the headset to the ground. Given this height estimate, we modify the height of the simulated human body to match that of the real human participant. 
We then used the VR headset to align the head and waist joints of a virtual human model to the person's pose.
At each time step, we queried the headset for its 3D Cartesian position, $\bm{\chi}$, and 3D orientation, $\bm{\theta}=(\theta_r, \theta_p, \theta_y)$.
We directly set the roll, pitch, and yaw joints of the simulated human head to align with the orientation, $\bm{\theta}$, of the headset. 

To align the waist pose, we computed the appropriate angles needed such that the simulated human had the same 3D Cartesian head position as the real person.
We first observed the fixed 3D position for the center of a person's waist, $\bm{W}$, given as a fixed offset above the wheelchair or hospital bed.
Let $\bm{\psi} = (\psi_x, \psi_y, \psi_z) = \bm{\chi} - \bm{W}$ be the 3D vector from the center of a person's waist to the center of their head.
We then computed the roll, pitch, and yaw $(r_{\bm{\psi}}, p_{\bm{\psi}}, y_{\bm{\psi}})$ orientations for the waist of the simulated human according to:
\begin{align*}
r_{\bm{\psi}} &= \text{atan2}(\psi_y, \psi_z)\\ 
p_{\bm{\psi}} &= \text{atan2}(\psi_x \cdot \text{cos}(r_{\bm{\psi}}), \psi_z)\\ 
y_{\bm{\psi}} &= \text{atan2}(\text{cos}(r_{\bm{\psi}}), \text{sin}(r_{\bm{\psi}})\cdot\text{sin}(p_{\bm{\psi}}))
\end{align*}
Finally, we observe that people often prefer to use some waist motion when looking left or right. To account for this, we distributed some of the measured head yaw orientation to the waist. Specifically, we set the yaw orientation of the head to be $0.7(\theta_y - y_{\bm{\psi}})$, and the yaw orientation of the waist joint to be $0.3(\theta_y - y_{\bm{\psi}})$.

As shown in Fig.~\ref{fig:vr}, participants also hold two Oculus Touch controllers, which we use to control both arms of the simulated human in virtual reality. At each time step, we queried each handheld controller for its 3D Cartesian position and orientation in global space.
Using inverse kinematics with a damped least squares method, we compute the 7-DoF arm joint angles necessary for the hand of the simulated person to have the same position and orientation as measured by the controller. 
Since we are optimizing for a 7-DoF arm pose using a 6-DoF controller pose measurement, this inverse problem is ill-posed, which implies that stable solutions are not guaranteed. However, we have found this inverse kinematics optimization to work well in practice, with only minor offsets in the estimated pose of a person's full arm.

\subsection{Test Environments}

\begin{figure}
\centering
\includegraphics[width=0.215\textwidth, trim={7cm 5cm 7cm 3cm}, clip]{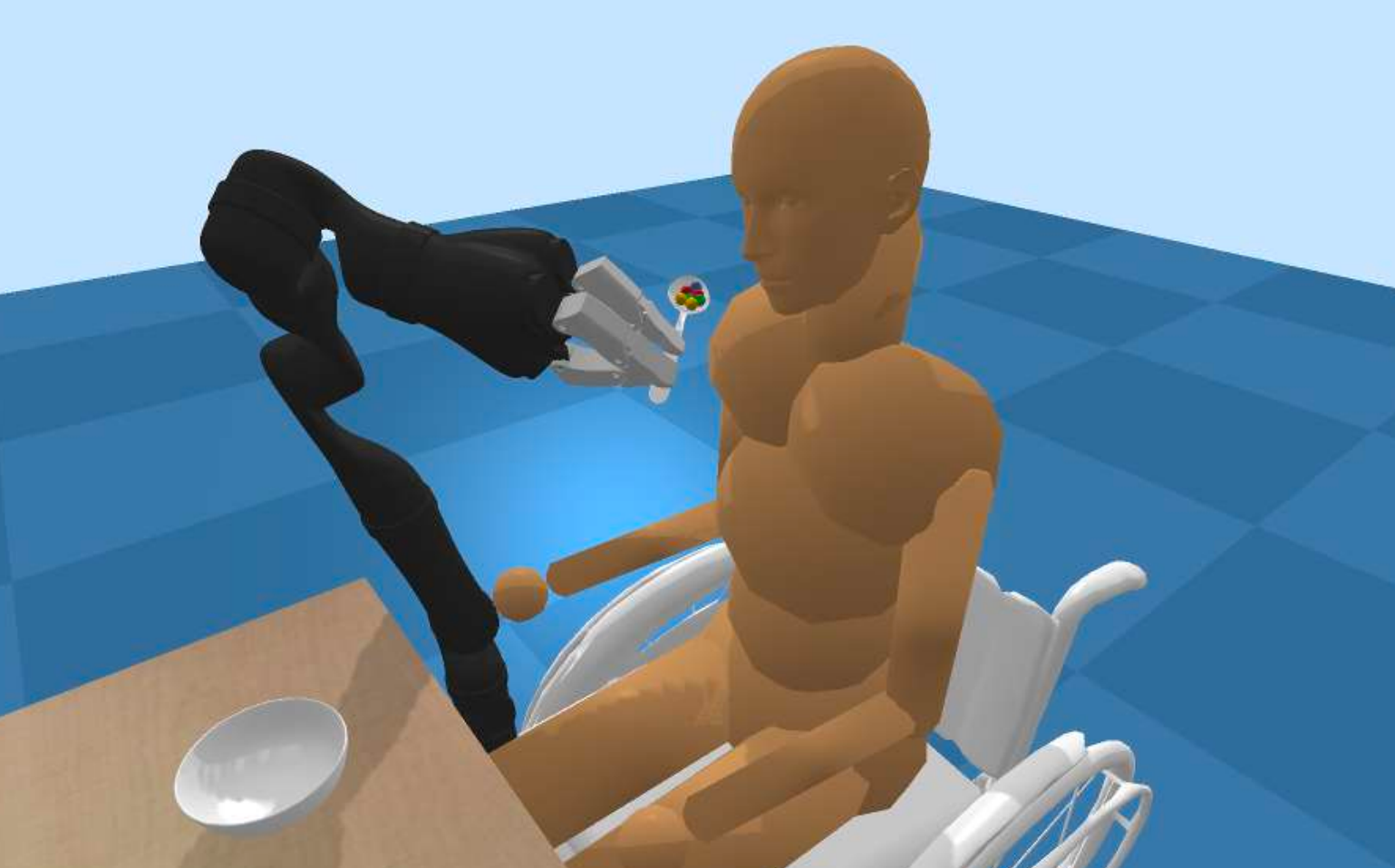}
\includegraphics[width=0.215\textwidth, trim={7cm 8cm 7cm 0cm}, clip]{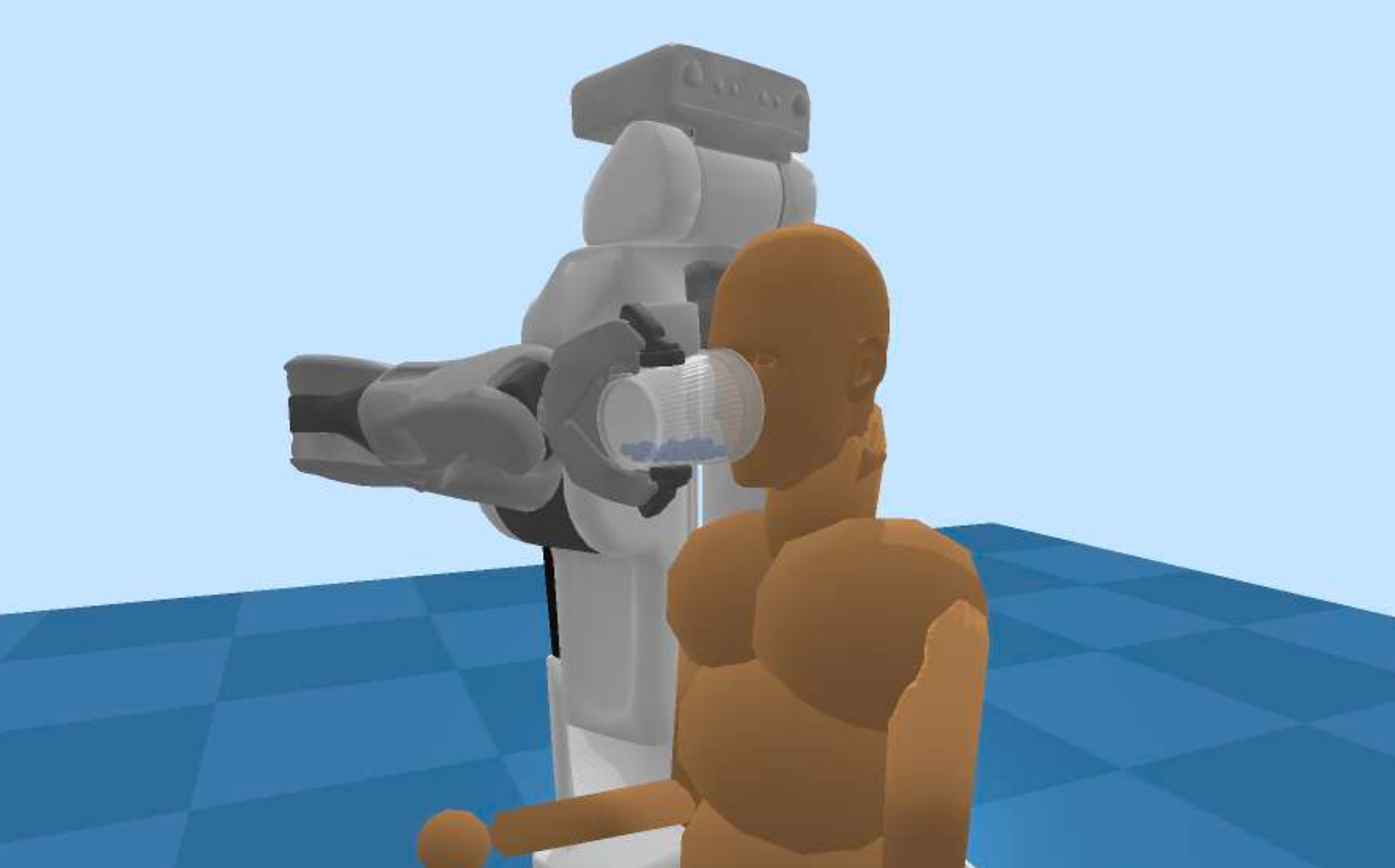}\\
\vspace{0.1cm}
\ \includegraphics[width=0.215\textwidth, trim={8.5cm 8.05cm 7.5cm 1.25cm}, clip]{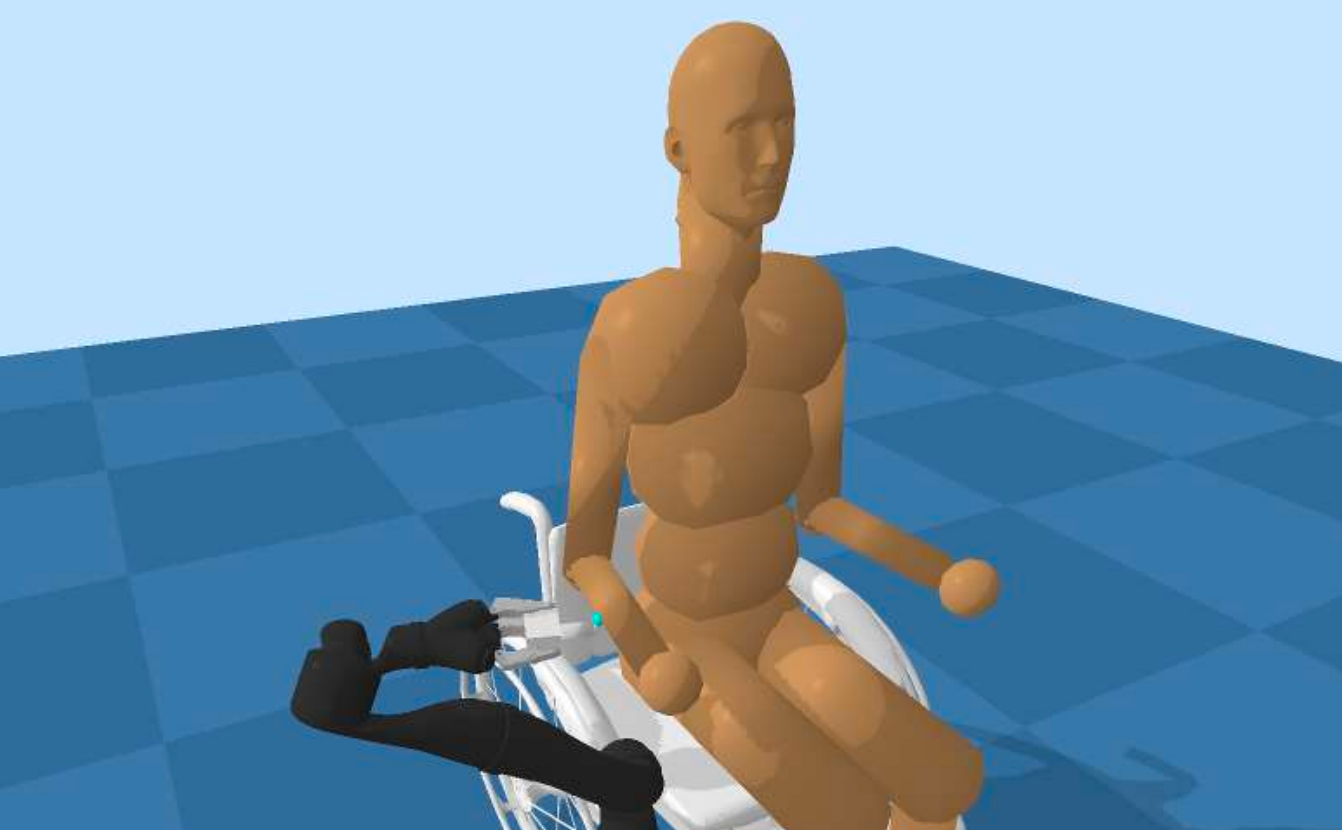}
\includegraphics[width=0.215\textwidth, trim={7cm 6.5cm 7cm 1.5cm}, clip]{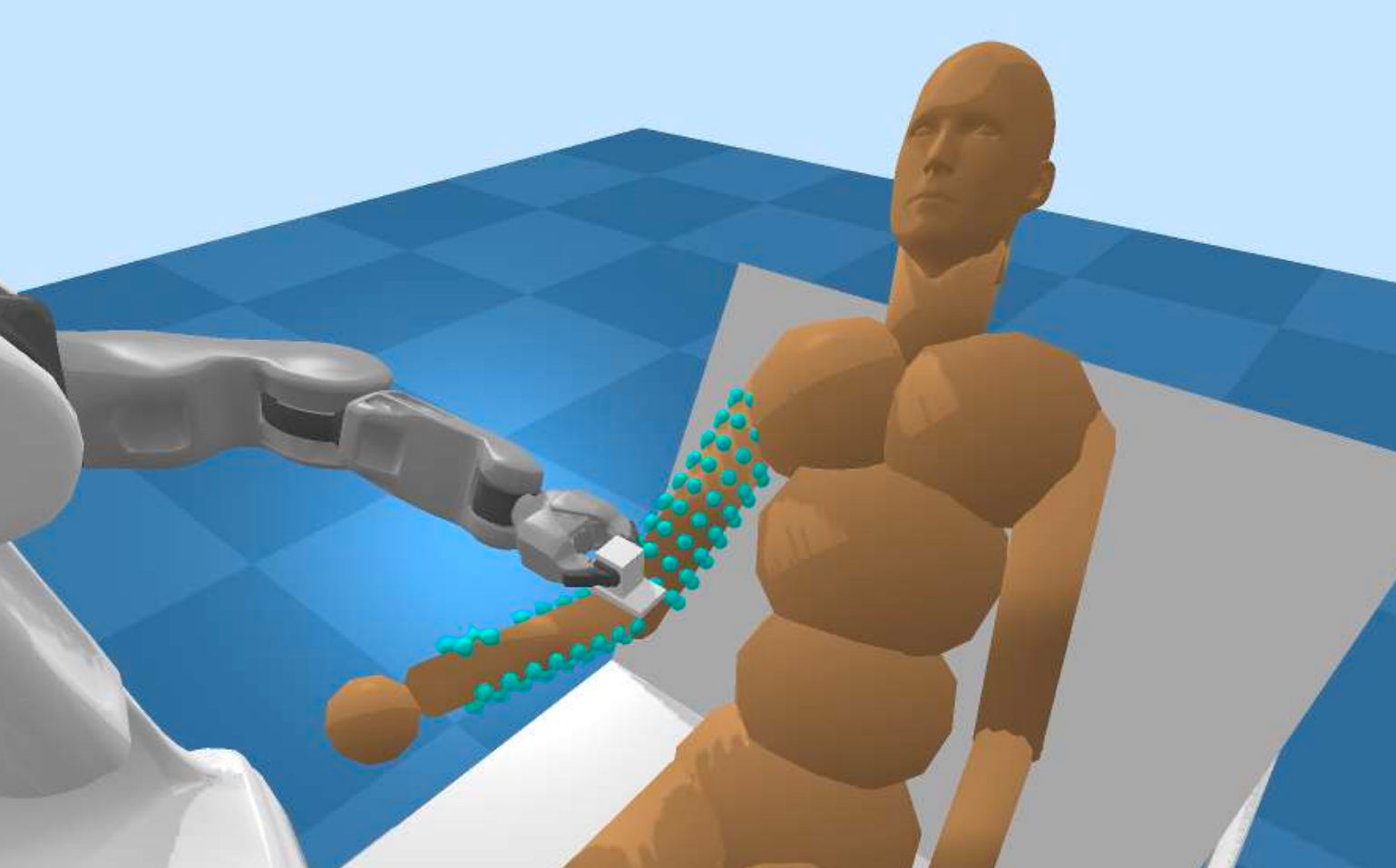}
\caption{\label{fig:envs}Assistive environments in physics simulation with the Jaco and PR2 robots. (Top row) Feeding and drinking assistance. (Bottom row) Itch scratching and bed bathing assistance.}
\vspace{-0.4cm}
\end{figure}

AVR Gym builds upon four physics-based assistive environments from Assistive Gym. A key aspect of achieving realism for users was to match the simulated wheelchair and bed to a real wheelchair and bed. 
Each of these environments, shown in Fig.~\ref{fig:envs}, is associated with activities of daily living (ADLs)~\cite{lawton1969assessment}, including:
\begin{itemize}[leftmargin=*]
\item \textbf{Feeding}: A robot holds a spoon and aims to move the spoon to a person's mouth. We place small spheres on the spoon, representing food. The person sits in a wheelchair during assistance. 
\item \textbf{Drinking Water}: A robot holds a cup containing small particles that represent water. The robot aims to move the cup toward the person's mouth and tilt the cup to help the person drink the water. The person sits in a wheelchair during assistance. 
\item \textbf{Itch Scratching}: A robot aims to scratch a randomly generated location along the person's right arm. The robot grasps a scratching tool in its left end effector. The person sits in a wheelchair during assistance. 
\item \textbf{Bed Bathing}: The robot holds a simulated washcloth tool while the person lies on a hospital bed in a randomly generated resting pose. The robot aims to move the washcloth around the person's right arm in order to clean the arm. 
\end{itemize}
These four assistive environments in AVR Gym are nearly identical to the original environments presented in~\cite{erickson2019assistive}, including the same male and female human models, realistic human joint limit models, robot base pose optimization, reward functions with human preferences, and task goals.

\section{Simulation-Trained Robot Control Policies}

Prior to developing AVR Gym, we trained eight robot control policies solely in simulation for the four caregiving tasks (robot-assisted feeding, drinking, itch scratching, and bed bathing) and two collaborative robots (the PR2 from Willow Garage and the 7-DoF Jaco (Gen2) arm from Kinova). These policies were the baseline policies released with Assistive Gym and described in the corresponding paper~\cite{erickson2019assistive}. We trained slightly modified versions of these eight robot control policies to run in AVR Gym, which we refer to as the \emph{Original Policies}. 

\subsection{Original Policy Training}
\label{sec:original_policies}

For each assistive task and robot, we follow the same training procedure presented in~\cite{erickson2019assistive}.
At each time step during simulation, the robot executes an action and then receives a reward and an observation based on the state of the world. Actions for a robot's 7-DoF arm are represented as changes in joint angles, $\Delta P \in \mathbb{R}^7$. The PR2 uses only its right or left arm depending on the assistive task. The observations for a robot include the 7D joint angles of the robot's arm, the 3D position and orientation of the end effector, and the forces applied at the end effector. The robot's observation also includes details of task-relevant human joints, including 3D positions of the shoulder, wrist, and elbow during the bed bathing and itch scratching assistive tasks, and the position and orientation of the person's head during feeding and drinking assistance. \new{Both the Jaco and PR2 robots use the same observation and reward functions during training.}

Our Original policies were trained using proximal policy optimization (PPO)~\cite{schulman2017proximal}, which is an actor-critic deep reinforcement learning approach that has recently shown success in assistive robotic contexts~\cite{cleggdissertation, clegg2020learning, erickson2019assistive}.
These policies are modeled using a fully-connected neural network with two hidden layers of 64 nodes and tanh activations. We train policies for 50,000 simulation rollouts (trials), where each rollout consists of 200 consecutive time steps (20 seconds of simulation time at 10 time steps per second).
Prior to each simulation rollout, we randomly initialize the simulated person's arm pose for itch scratching and bed bathing tasks, and we randomize the head orientation for feeding and drinking tasks. Once initialized, the human holds a static pose throughout the entire rollout. Each policy is trained with default male and female human models, with heights, body sizes, weights, and joint limits matching published 50th percentile values~\cite{tilley2002measure}. Note that these control policies are not temporal models and do not consider observations from previous time steps in a rollout. As such, these models are not affected by the specific nature of human motion, allowing us to evaluate these policies with real people without needing to model realistic human motion in simulation.

\subsection{Revised Policies}
\label{sec:improved_policies}

During early pilot studies in AVR Gym with lab members, we observed that the Original policies exhibited unexpected deficiencies and poor performance when providing assistance to human participants in virtual reality.
Most notable were failures in the itch scratching and bed bathing assistance tasks, where the robots would fail to move their end effectors closer to the person's body. Through iterative investigation and development with real people in AVR Gym, we developed the \emph{Revised Policies}, based on the key discovery that the biomechanical models of simulated people we used to train the Original policies significantly differed from the biomechanics of real people. The two leading factors were variation among human heights and waist bending movements, wherein the Original policies were trained on male and female models with fixed heights and were not trained on any variation among the human waist joints.

Given these findings, we modified each simulation environment such that the simulated human biomechanics better match people in VR. Specifically, we randomized the simulated human torso height and initial waist orientation before each training trial began.
The person's torso height, measured from hipbone to center of the head, was uniformly sampled from 50~cm to 70~cm for male models and 44~cm to 64~cm for female models. We then randomized the simulated person's initial three kinematic waist joints to angles within the range of (-10$^{\circ}$, 10$^{\circ}$). With these improved biomechanical simulations, we trained a new set of eight \emph{Revised Policies}, one for each robot and assistive task, using the same training process discussed in Section~\ref{sec:original_policies}.

When evaluated with simulation humans, these Revised policies achieve similar rewards and task success rates as the Original policies. However, as confirmed by our formal study (Section~\ref{sec:evaluation}), the Revised policies overcame the limitations of the Original policies and performed significantly better across both subjective and objective metrics when evaluated with real people in VR.

\begin{figure*}
\centering
\includegraphics[width=0.24\textwidth, trim={2cm 2cm 1cm 1cm}, clip]{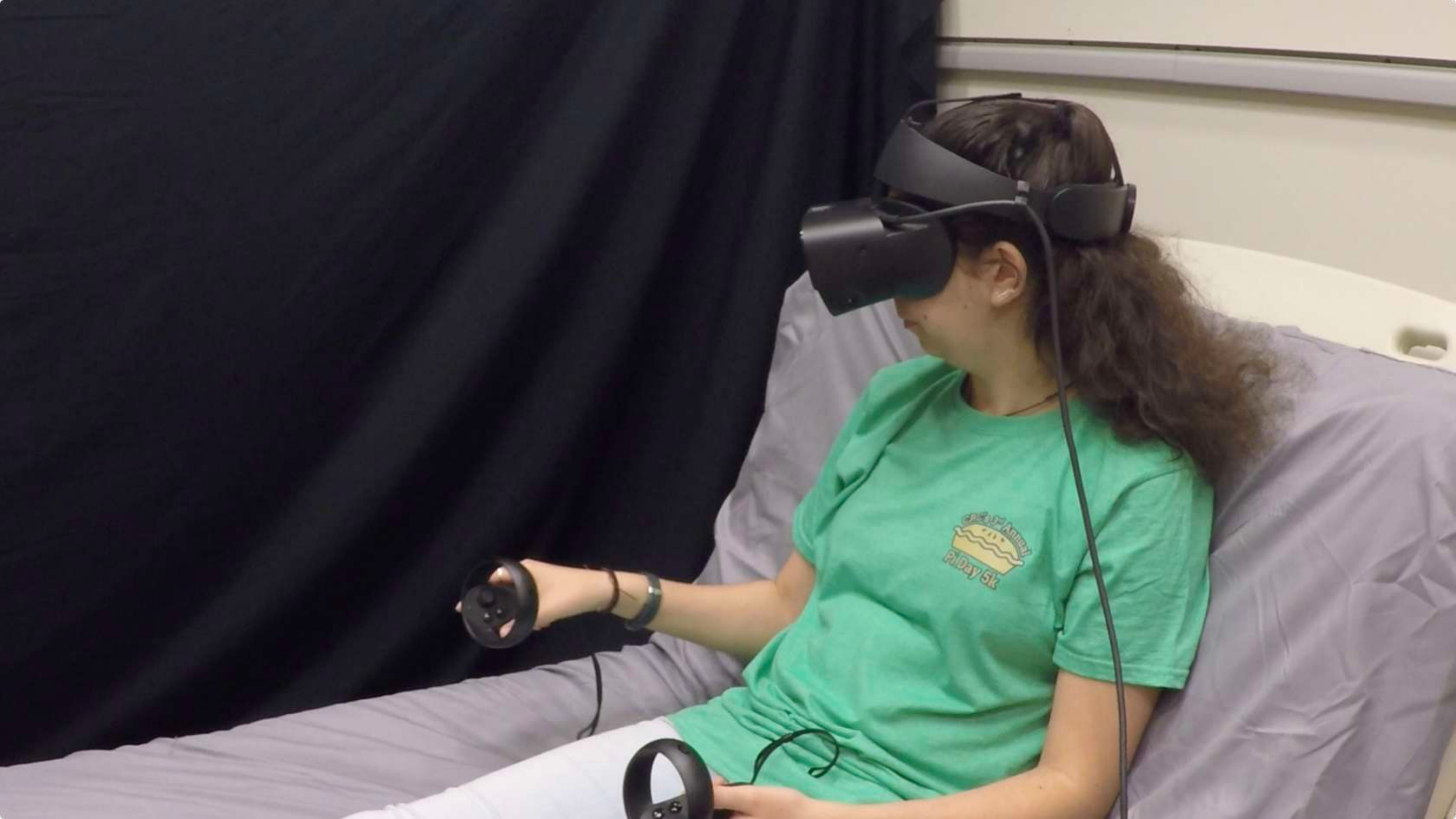}
\includegraphics[width=0.24\textwidth, trim={2cm 2cm 1cm 1cm}, clip]{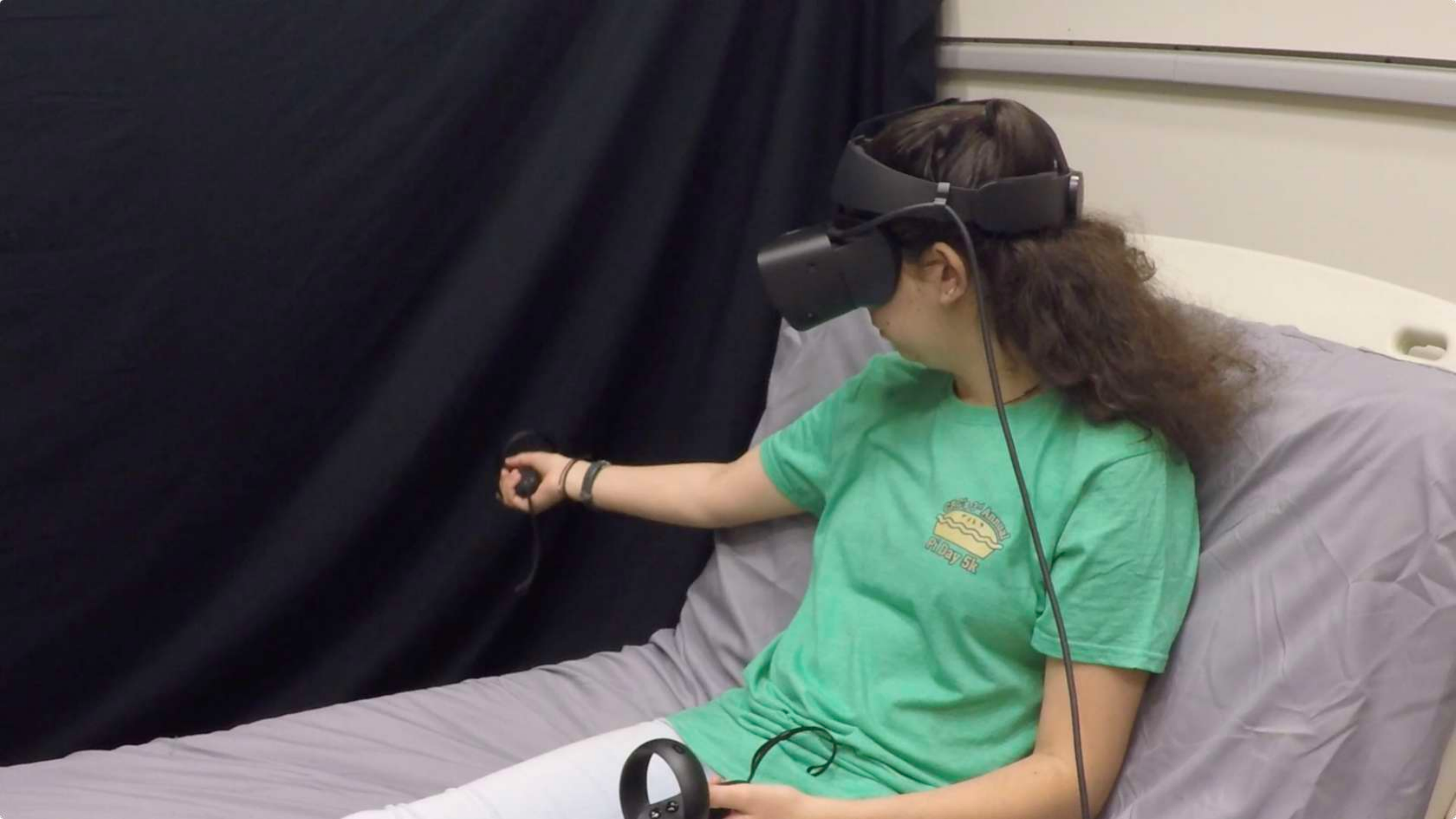}
\includegraphics[width=0.24\textwidth, trim={2cm 2cm 1cm 1cm}, clip]{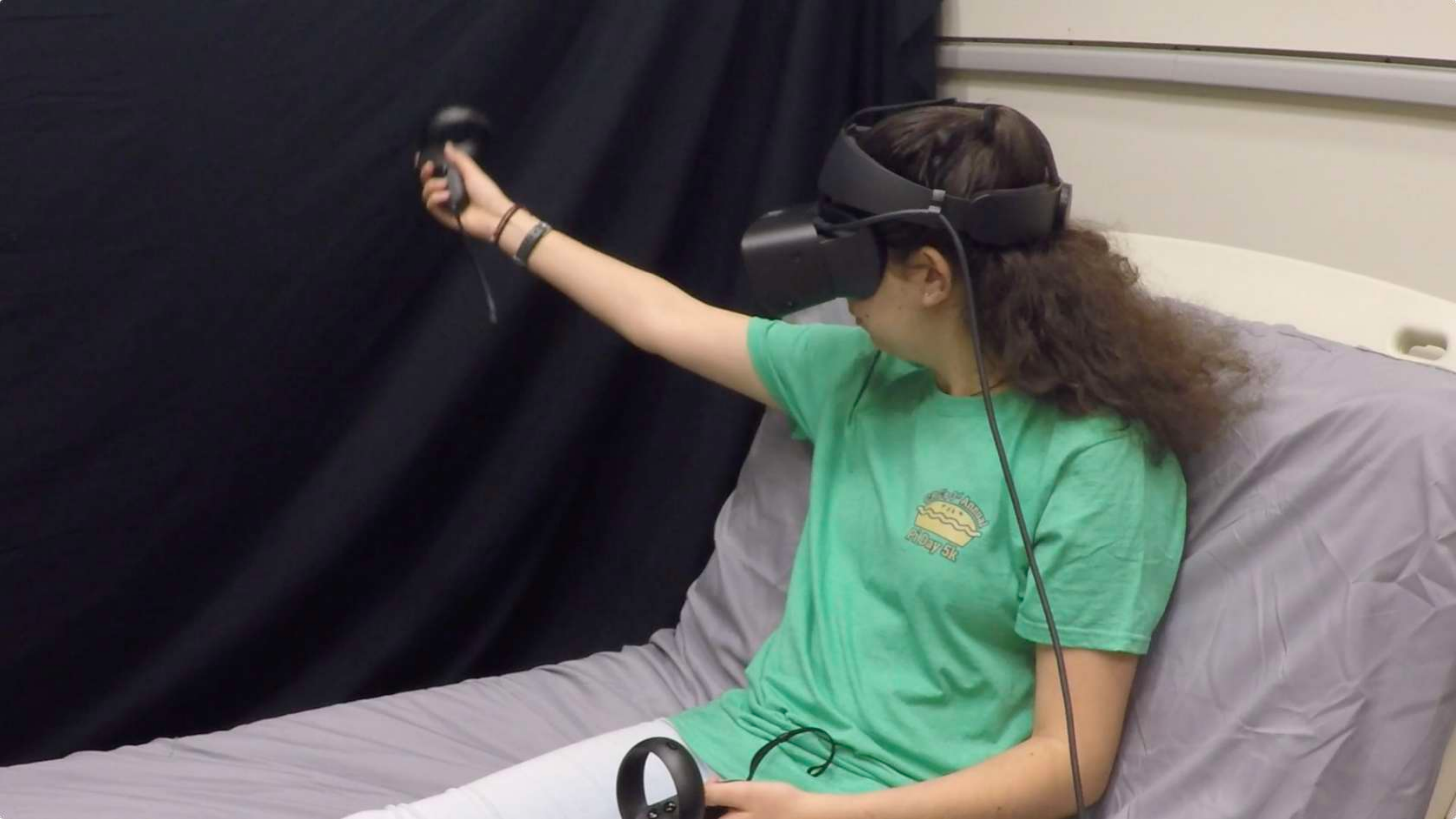}
\includegraphics[width=0.24\textwidth, trim={2cm 2cm 1cm 1cm}, clip]{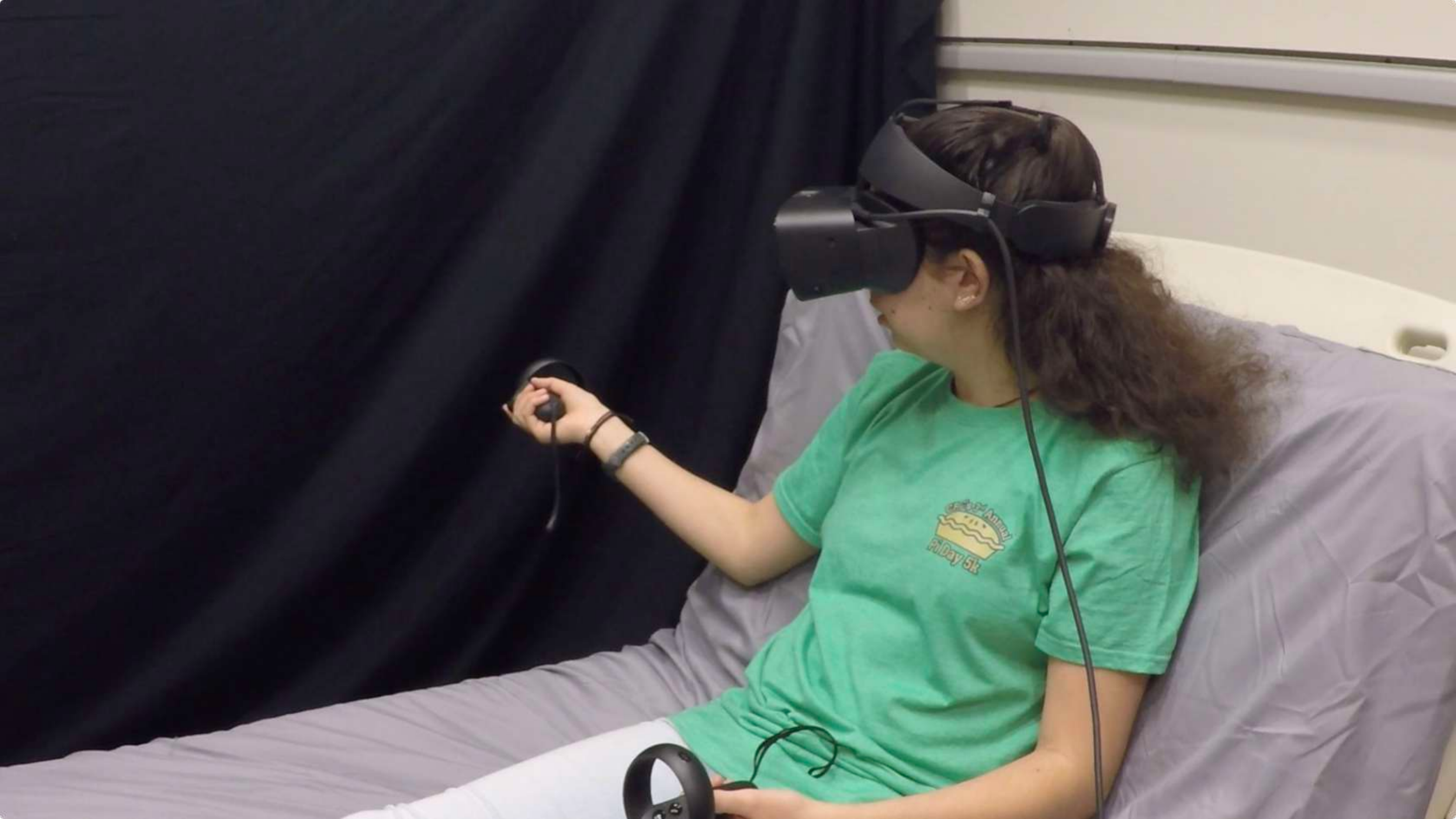}\\
\ \includegraphics[width=0.24\textwidth, trim={2cm 2cm 1cm 1cm}, clip]{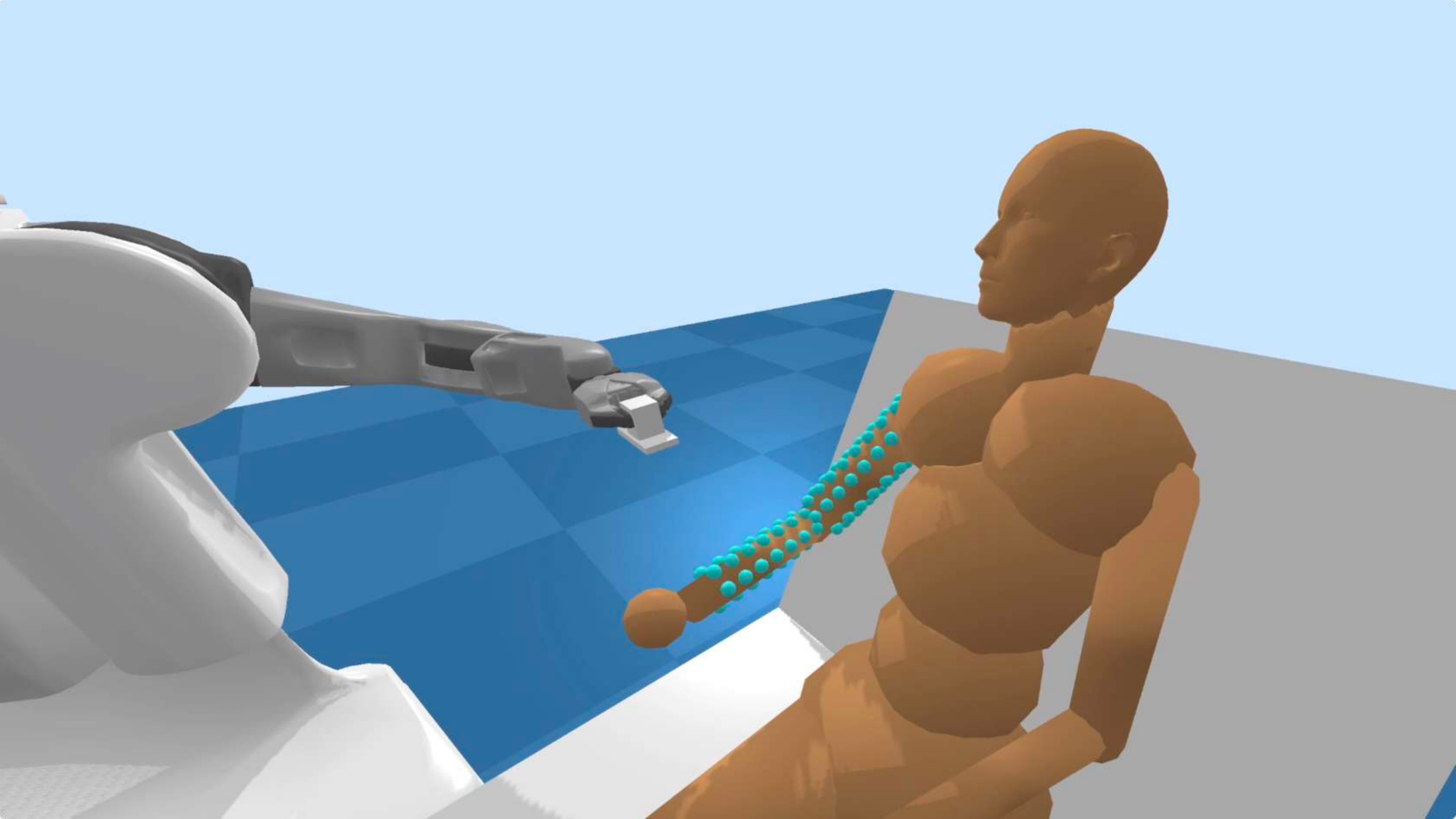}
\includegraphics[width=0.24\textwidth, trim={2cm 2cm 1cm 1cm}, clip]{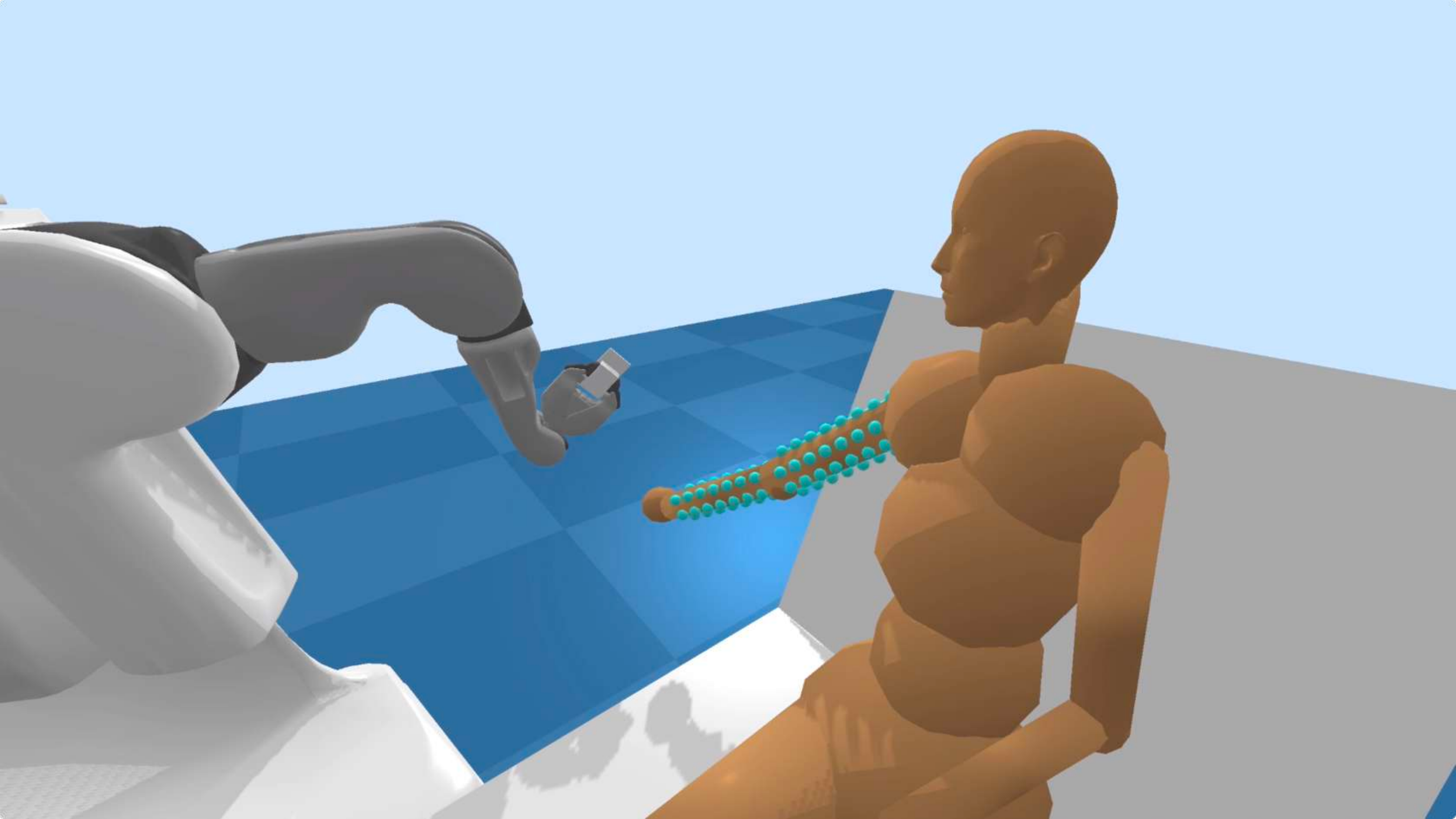}
\includegraphics[width=0.24\textwidth, trim={2cm 2cm 1cm 1cm}, clip]{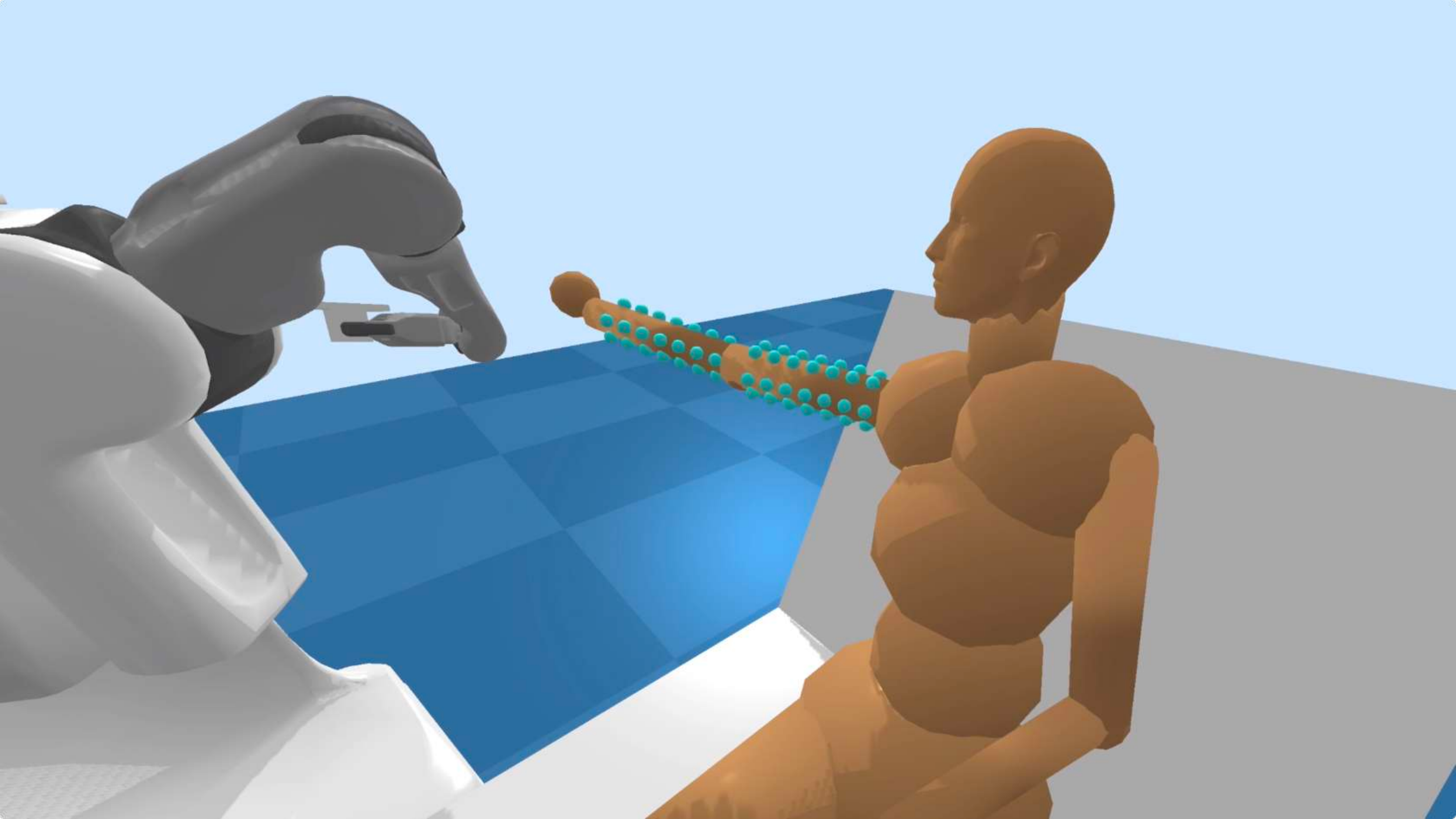}
\includegraphics[width=0.24\textwidth, trim={2cm 2cm 1cm 1cm}, clip]{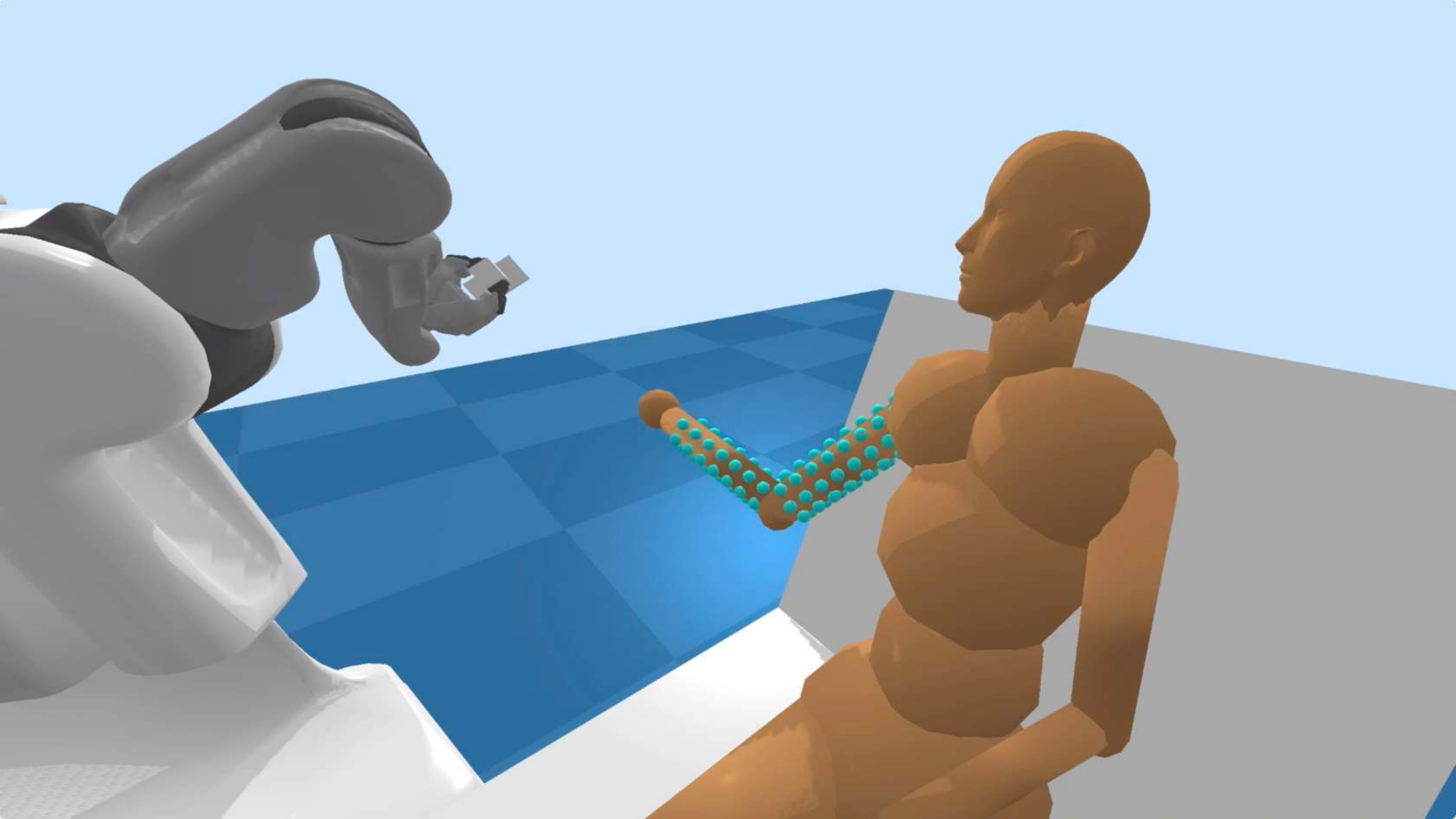}
\vspace{-0.2cm}
\caption{\label{fig:bed_bathing_fail}The PR2 robot fails to provide assistance with bed bathing in virtual reality when using the Original control policy. Blue markers are placed uniformly around the person's right arm, which the robot can clean off with the bottom of the wiping tool.}
\vspace{-0.2cm}
\end{figure*}

\begin{figure*}
\centering
\includegraphics[width=0.24\textwidth, trim={1cm 3cm 3cm 1cm}, clip]{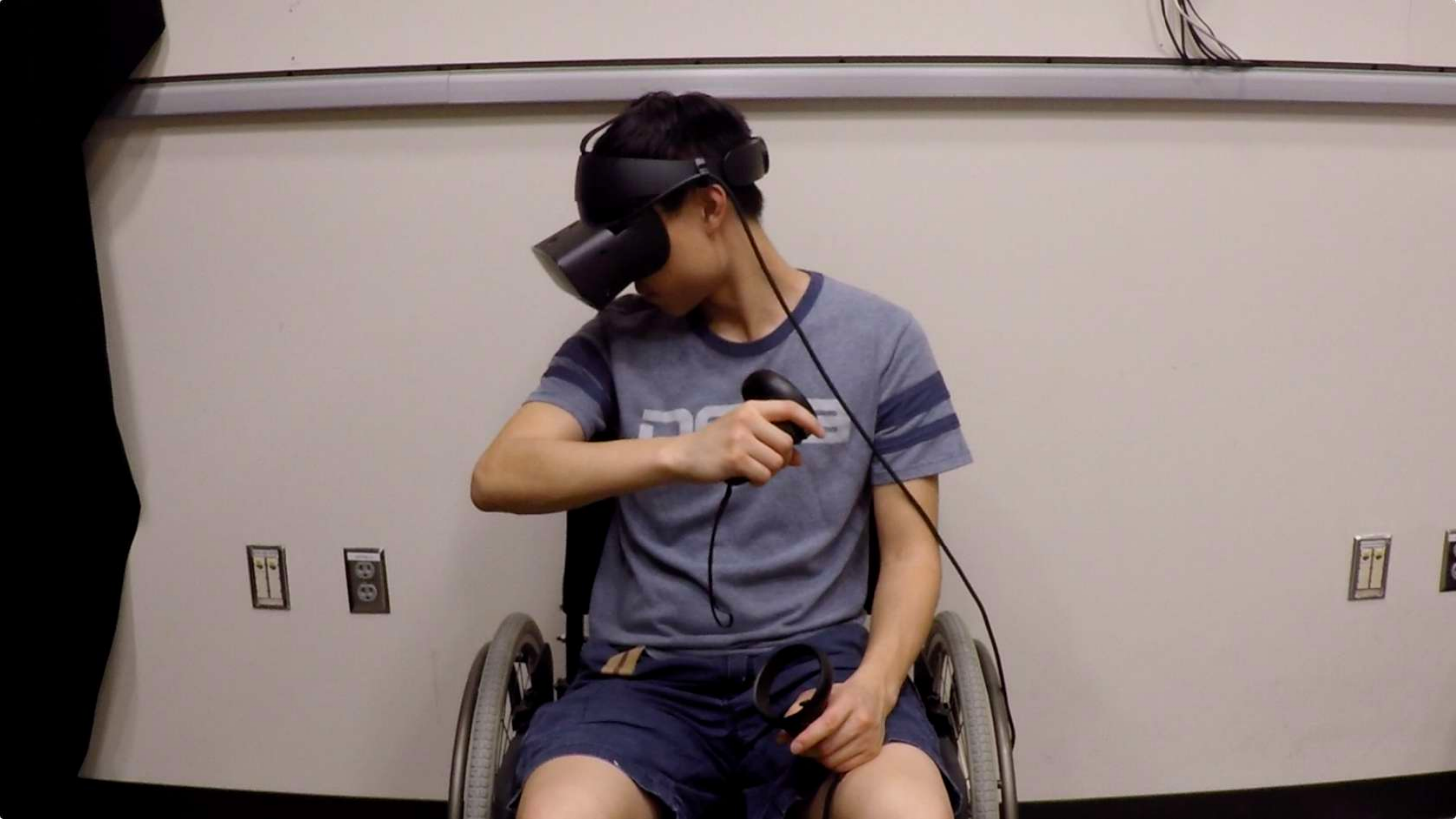}
\includegraphics[width=0.24\textwidth, trim={1cm 3cm 3cm 1cm}, clip]{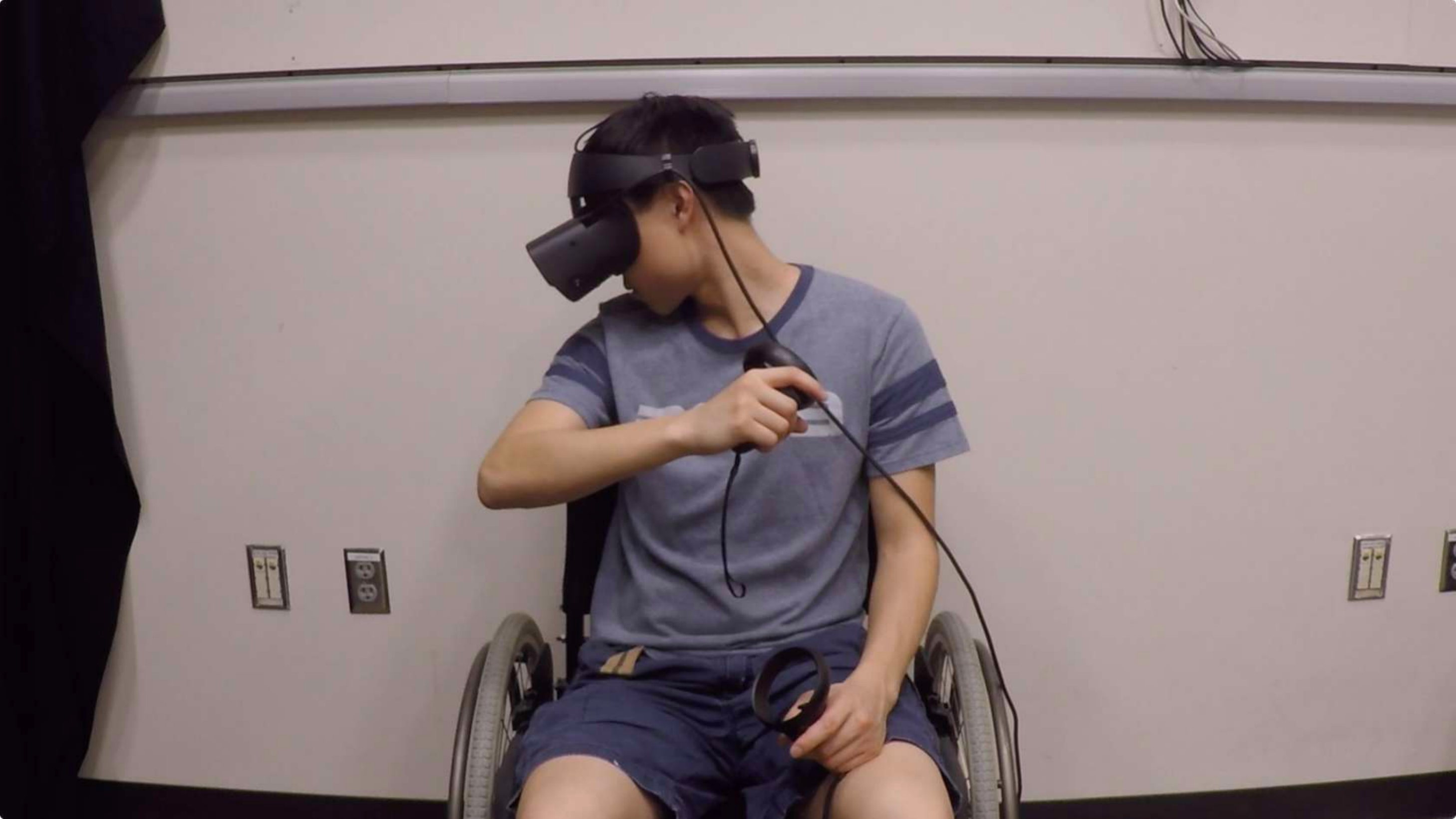}
\includegraphics[width=0.24\textwidth, trim={1cm 3cm 3cm 1cm}, clip]{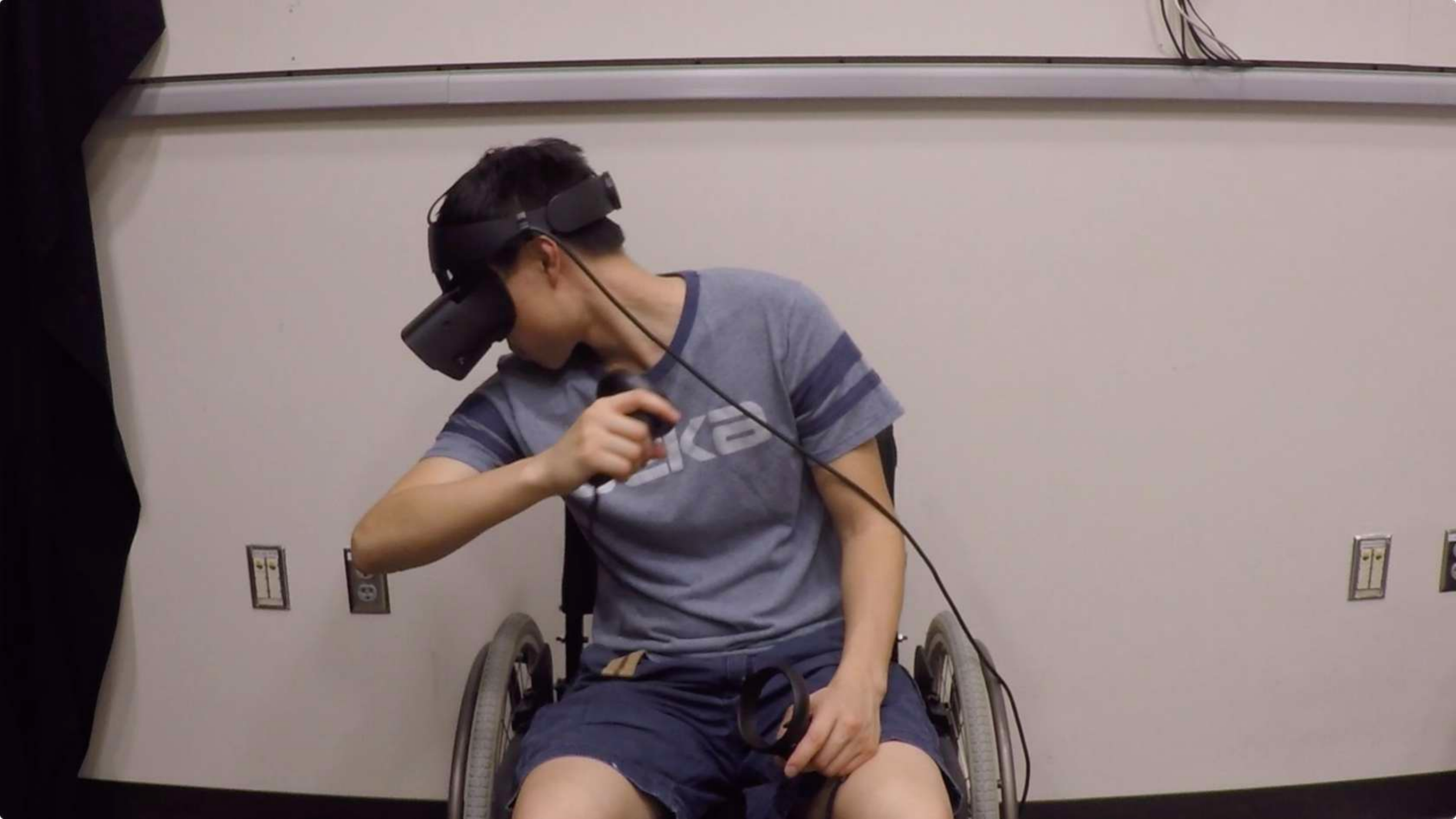}
\includegraphics[width=0.24\textwidth, trim={1cm 3cm 3cm 1cm}, clip]{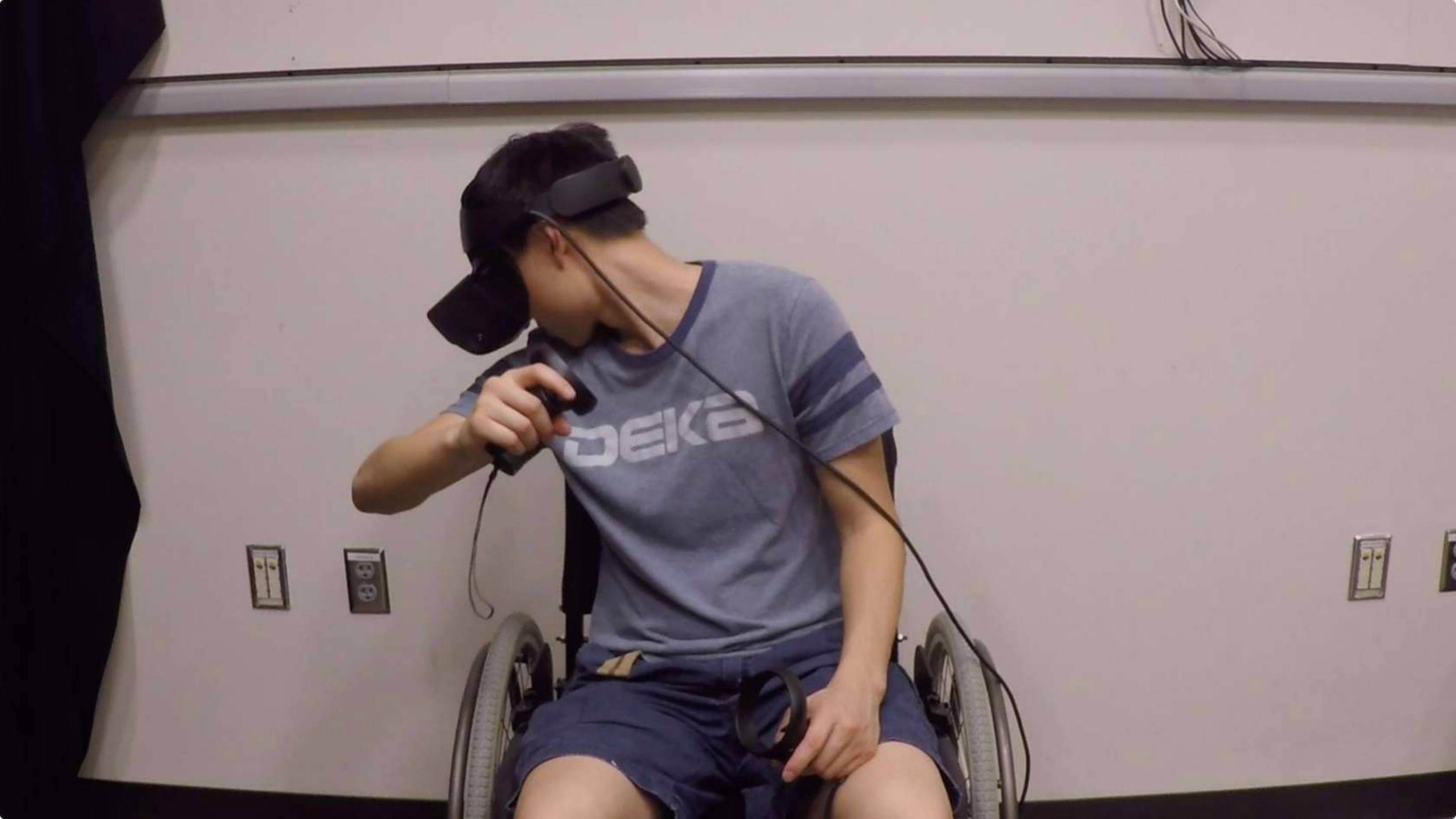}\\
\ \includegraphics[width=0.24\textwidth, trim={1cm 4cm 3cm 0cm}, clip]{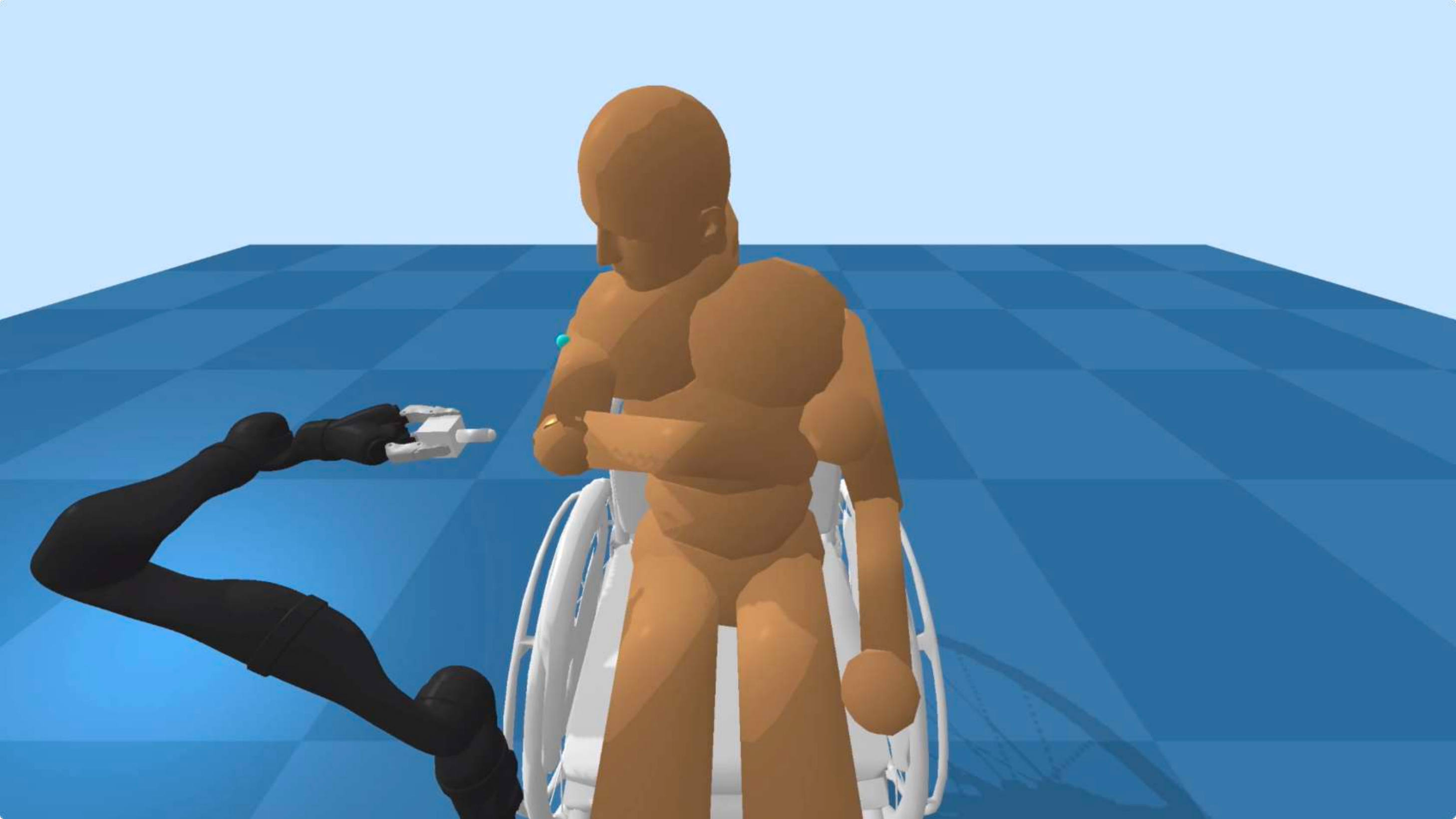}
\includegraphics[width=0.24\textwidth, trim={1cm 4cm 3cm 0cm}, clip]{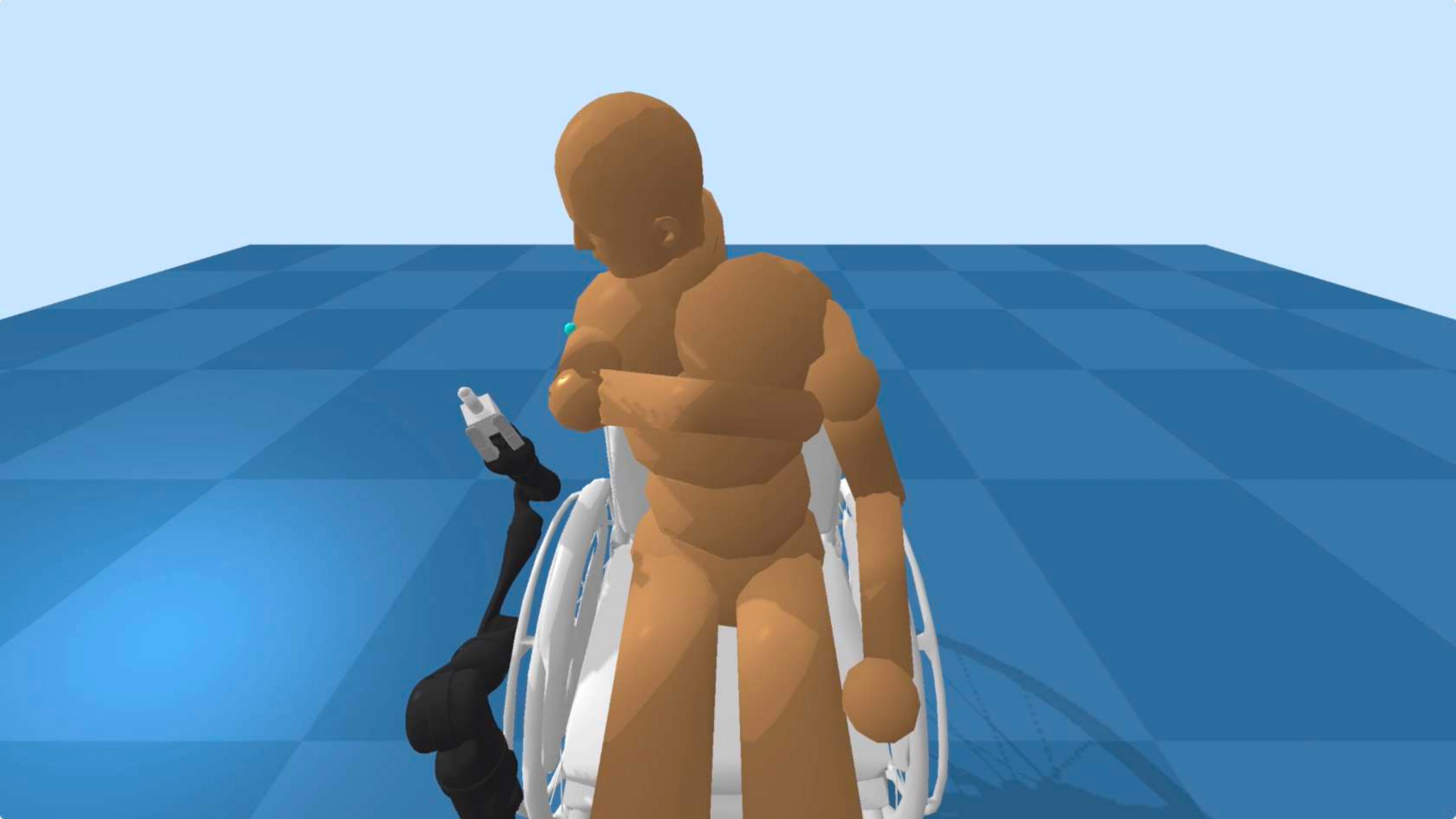}
\includegraphics[width=0.24\textwidth, trim={1cm 4cm 3cm 0cm}, clip]{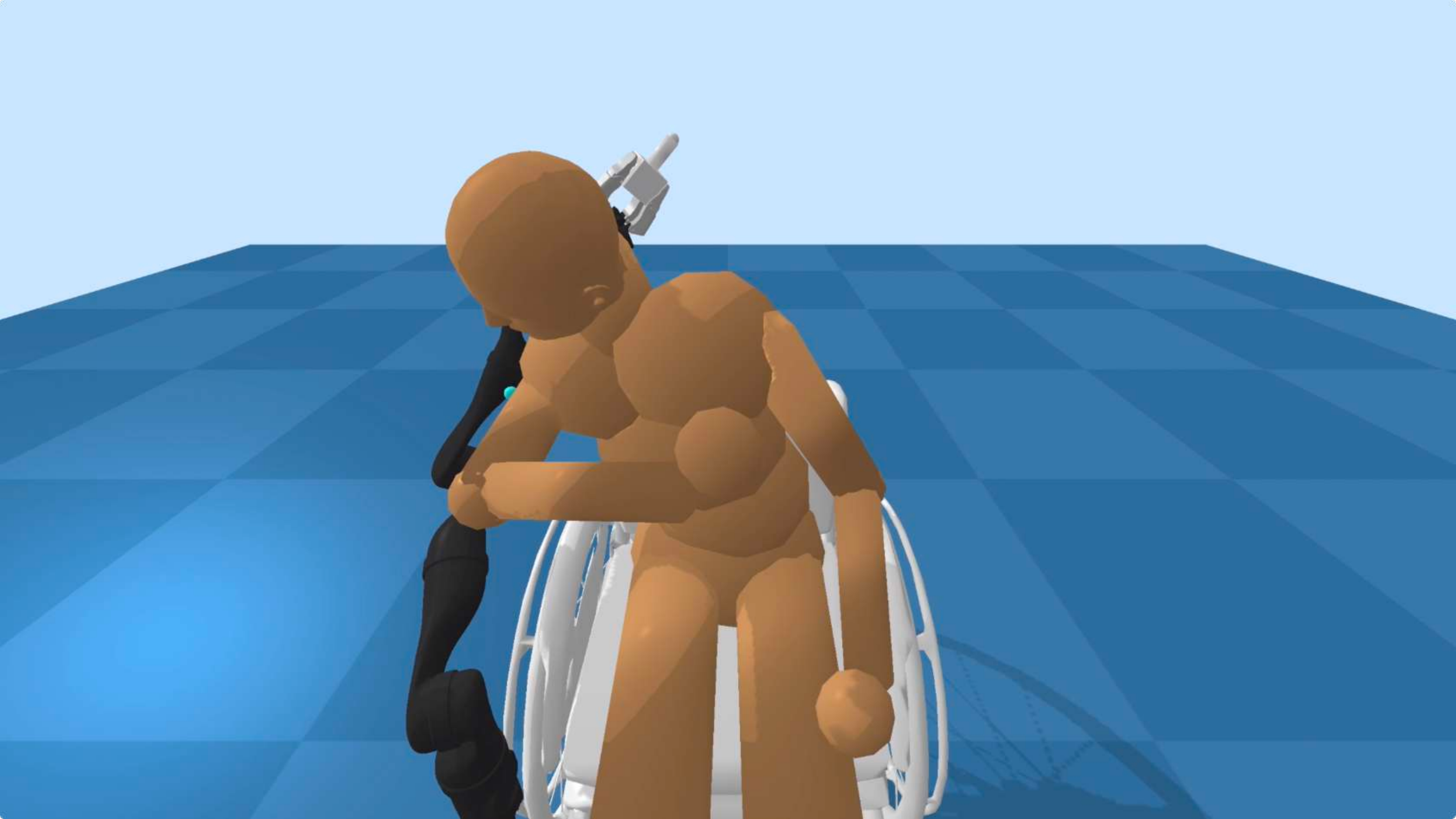}
\includegraphics[width=0.24\textwidth, trim={1cm 4cm 3cm 0cm}, clip]{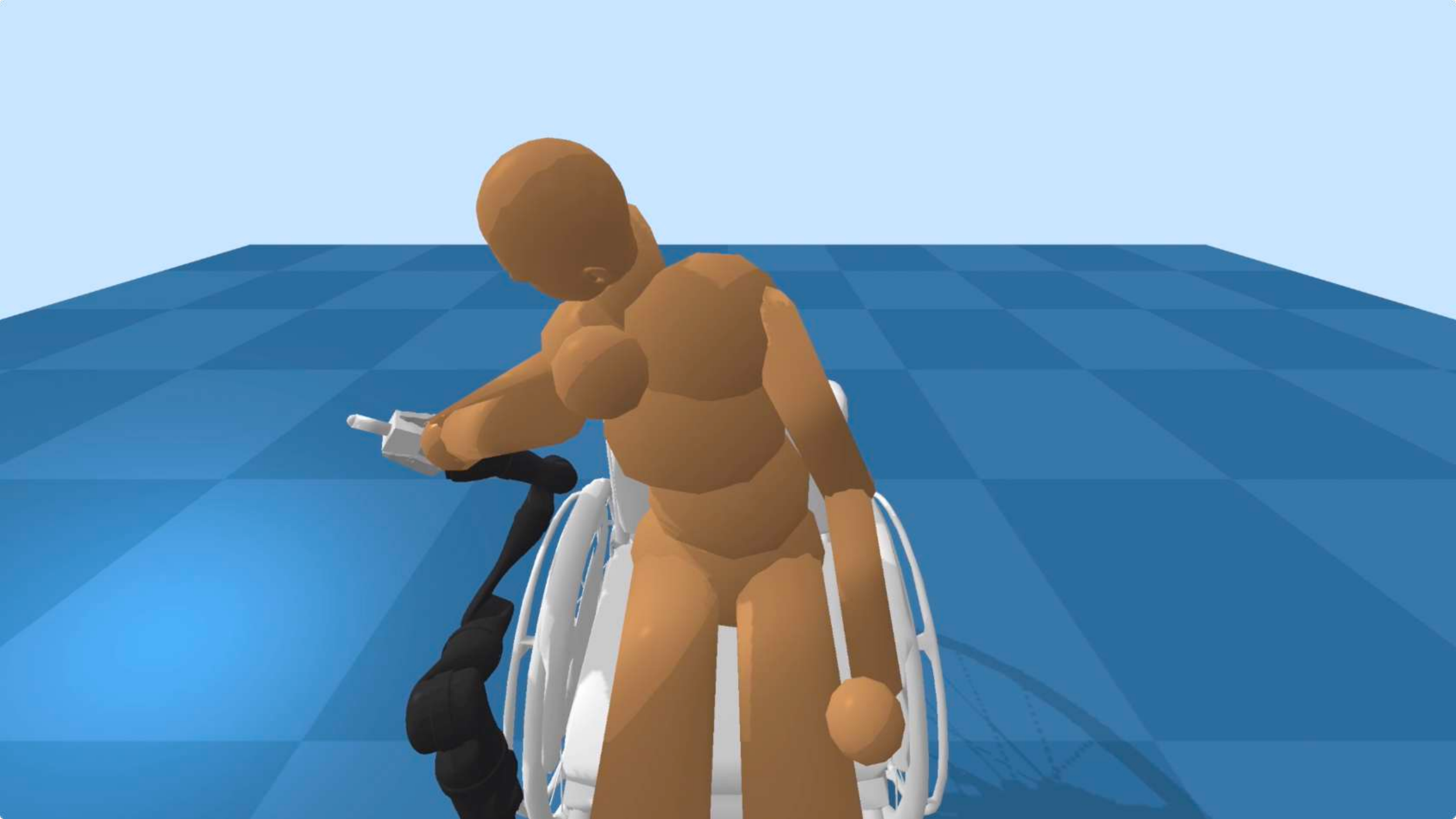}
\vspace{-0.2cm}
\caption{\label{fig:itch_scratching_fail}The Jaco fails to provide itch scratching assistance in virtual reality with the Original control policy. The itch is represented by a blue marker.}
\vspace{-0.3cm}
\end{figure*}

\section{Formal Study}
\label{sec:evaluation}

We conducted a formal study with eight participants to evaluate the performance of the Original policies and the Revised policies in terms of objective and subjective measures. In addition, we performed a posthoc analysis of biomechanical differences between the simulated humans used to train the Original policies, the simulated humans used to train the Revised policies, and the real humans who participated in our study. 

\subsection{Experimental Procedure}

We conducted an experiment with eight able-bodied human participants (four females and four males). We obtained informed consent from all participants and approval from the Georgia Institute of Technology Institutional Review Board (IRB). We recruited participants to meet the following inclusion/exclusion criteria: 18 years of age or older; able-bodied; no cognitive or visual impairments; fluent in spoken and written English; and not diagnosed with epilepsy. Participant ages ranged from 19 to 24 years old with torso heights varied between 0.51 and 0.58 meters.

For each participant, we conducted a total of 16 trials in VR with four trials for each of the four assistive tasks. The four trials for each task were organized such that we evaluated both the Original and Revised policies on both the PR2 and Jaco robots. We randomized the ordering of assistive tasks, robots, and control policies for each participant according to a randomized counterbalancing design.

During the study, participants wore the virtual reality headset and held both controllers. Participants sat in a wheelchair for the itch scratching, feeding, and drinking tasks and laid in a hospital bed for bed bathing assistance.
Similar to the simulation environments used for training control policies, each trial in virtual reality lasted 20 seconds for a total of 200 time steps. The robot executed an action at each time step in virtual reality, once every 0.1 seconds. 
Participants were instructed that they could move their arms, upper body, and head to interact with the simulated robots towards the goal of successfully receiving assistance for each task.
Prior to each assistive task, we gave participants an unscripted practice trial to familiarize themselves on interacting with the robot and how to accomplish the task in virtual reality. We randomly selected either the PR2 or Jaco robot and used the Revised control policies (see Section~\ref{sec:improved_policies}) for each practice trial. 
In order to elicit a wide range of human motion and interactions throughout our study, we instructed participants to accomplish a task (e.g. drink water from the cup), but we did not provide instructions on how to interact with the robot or how to appropriately complete the task.
Our study and experiments can be found in the supplementary video.

\subsection{Objective Measures}

We used the reward functions and success percentages defined in Assistive Gym as objective measures of performance~\cite{erickson2019assistive}.

\begin{itemize}[leftmargin=*]
\item \textbf{Feeding}: The robot is rewarded for moving the spoon closer to the person's mouth and when food enters the person's mouth. The robot is penalized for dropping food or applying large forces to the person. Task success is defined by the robot feeding at least 75\% of all food particles to the person's mouth.
\item \textbf{Drinking Water}: The robot is rewarded for moving the cup closer to the person's mouth, tilting the cup for drinking, and when water particles enter the person's mouth. The robot is penalized for spilling water or applying large forces to the person. Task success is defined by the robot pouring at least 75\% of all water into the person's mouth.
\item \textbf{Itch Scratching}: The robot is rewarded for moving the scratching tool closer to the itch location and for performing scratching motions around the itch. The robot is penalized for applying large forces to the person, or more than 10~N of force~\cite{erickson2018deep} near the itch. Task success is defined by the robot performing at least 25 scratching motions near the itch. 
\item \textbf{Bed Bathing}: The robot is rewarded for moving the washcloth closer to the person's arm and for using the bottom of the washcloth to wipe off markers uniformly distributed (3~cm apart) along the person's right arm. The robot is penalized for applying large forces to the person. Task success is defined by the robot wiping off at least 30\% of all markers along a person's arm)
\end{itemize}

\subsection{Subjective Measures}

In order to assess participants' perceptions of the control policies, we used a questionnaire with four statements pertaining to perceptions of the robot's performance, safety, comfort and speed. For each statement, participants were asked to record how much they agreed with the statement on an interval scale from 1 (strongly disagree) to 7 (strongly agree) with 4 being neutral. We based this 7-point scale questionnaire on past work on robot-assisted feeding~\cite{park2020active}.  

The four statements follow:
\begin{itemize}
    \item L1: The robot successfully assisted me with the task.
    \item L2: I felt comfortable with how the robot assisted me in VR.
    \item L3: I felt I would be safe if this were a real robot.
    \item L4: The robot moved with appropriate speed.
\end{itemize}

\section{Results}
\subsection{Performance of the Original Policies}
\label{sec:hypothesis}

\begin{table}
\centering
\caption{\label{table:errors_original}Average reward and task success between simulation and virtual reality using the Original control policies.}
\begin{tabular}{cccccc} \toprule
    & \multicolumn{2}{c}{Simulation} & \multicolumn{2}{c}{Virtual Reality} \\ \cmidrule{2-5}
    Task & PR2 & Jaco & PR2 & Jaco \\ \midrule\midrule
    Feeding & 103 (77\%) & 82 (75\%) & 122 (100\%) & 37 (63\%) \\
    Drinking & 380 (59\%) & 144 (32\%) & 502 (75\%) & -92 (0\%) \\
    Scratching & 62 (35\%) & 218 (48\%) & 25 (25\%) & -69 (0\%) \\
    Bathing & 85 (13\%) & 118 (28\%) & -87 (0\%) & -126 (0\%) \\
    \midrule 
    Avg. Success & 46\% & 46\% & 50\% & 16\% \\
	\bottomrule
\end{tabular}
\vspace{-0.4cm}
\end{table}

Table~\ref{table:errors_original} depicts the average reward and task success when both robots used the Original policies in simulation with static human models and in AVR Gym with the eight human participants. 
When evaluating the Original policy for a given robot and assistive task in virtual reality, we averaged rewards and task success over trials from all eight participants. For simulation, we averaged rewards and task success over 100 simulation rollouts with a random initial human pose for each rollout.
\new{Since the simulated human holds a static pose throughout an entire trial, we take an average over a larger number of simulation trials to evaluate performance over multiple human poses.} \new{We note that each task uses a slightly different reward function due to task-specific reward elements and hence reward values are not directly comparable across different tasks.}

For the feeding assistance task, the Original policies achieved similar average reward and task success with real people as they achieved in simulation with fixed human models.
However, performance was more varied for the drinking, itch scratching, and bed bathing tasks.
The most noticeable errors occurred with the itch scratching and bed bathing tasks, where the Original policies for both the Jaco and PR2 robots frequently failed to move their end effectors closer to a person's body.
These errors can be visually seen in Fig.~\ref{fig:bed_bathing_fail} where the PR2 actively moved away from the person during bed bathing assistance and in Fig.~\ref{fig:itch_scratching_fail} where the Jaco robot actuated itself to behind the wheelchair rather than scratching an itch on the person's arm.
\new{Differences in results between the two robots can be partially attributed to differences in robot kinematics and base positioning. The Jaco is mounted to a fixed position on the wheelchair, whereas we optimize the PR2's base pose near a person according to joint-limit-weighted kinematic isotropy (JLWKI) and task-centric manipulability metrics~\cite{erickson2019assistive, kapusta2019task}.}

Except for responses about the robot's speed (L4), participants' average responses were negative or neutral (L1, L2, and L3) (see Fig.~\ref{fig:likert}). Most notably, participants tended to perceive the robots as being unsuccessful at assistance, as evidenced by responses to L1 being significantly below neutral (4) with $p<0.05$ using a Wilcoxon signed-rank test.

Responses did vary by task. When the robots used the Original policies for itch scratching assistance, participants slightly disagreed with the statement on successful assistance (L1), with an average response of 3.3 across all participants. When using the Original policies for bed bathing assistance, participants reported an even lower average rating of 1.2, where almost all participants strongly disagreed that the robots provided successful bathing assistance. 

\subsection{Performance of the Revised Policies}
\label{sec:improvedeval}

\begin{table}
\centering
\caption{\label{table:original_improved_compare}Average reward and task success for the Original and Revised control policies in VR with real people.}
\begin{tabular}{cccccc} \toprule
    & \multicolumn{2}{c}{Original Policies} & \multicolumn{2}{c}{Revised Policies} \\ \cmidrule{2-5}
    Task & PR2 & Jaco & PR2 & Jaco \\ \midrule\midrule
    Feeding & 122 (100\%) & 37 (63\%) & 113 (100\%) & 36 (45\%) \\
    Drinking & 502 (75\%) & -92 (0\%) & 458 (75\%) & 199 (42\%) \\
    Scratching & 25 (25\%) & -69 (0\%) & 36 (45\%) & 95 (62\%) \\
    Bathing & -87 (0\%) & -126 (0\%) & -18 (0\%) & -16 (0\%) \\
    \midrule 
    Avg. Success & 50\% & 16\% & \textbf{55\%} & \textbf{37\%} \\
	\bottomrule
\end{tabular}
\vspace{-0.5cm}
\end{table}

\begin{figure*}
\centering
\includegraphics[width=0.24\textwidth, trim={3cm 3cm 3cm 2cm}, clip]{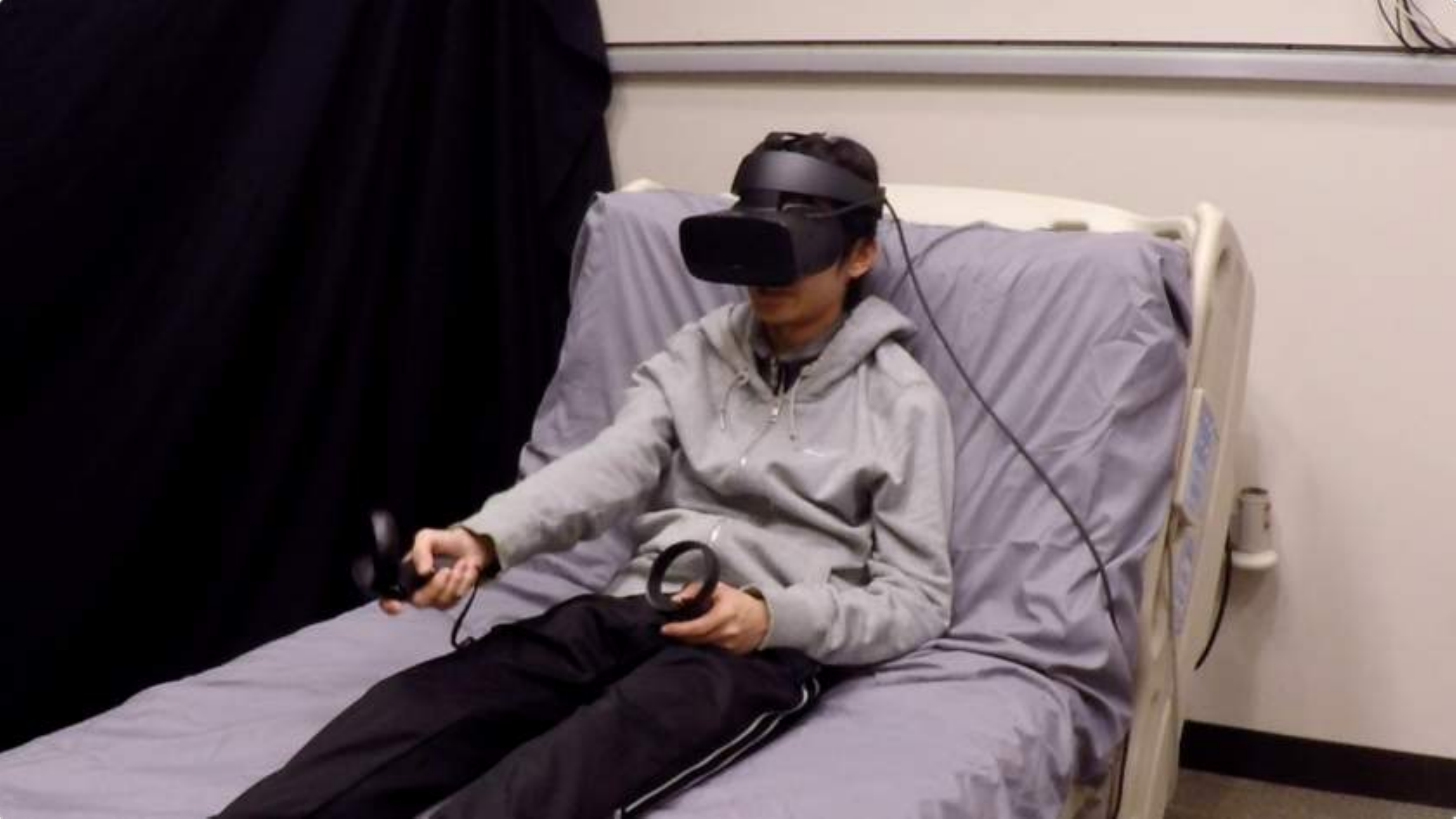}
\includegraphics[width=0.24\textwidth, trim={3cm 3cm 3cm 2cm}, clip]{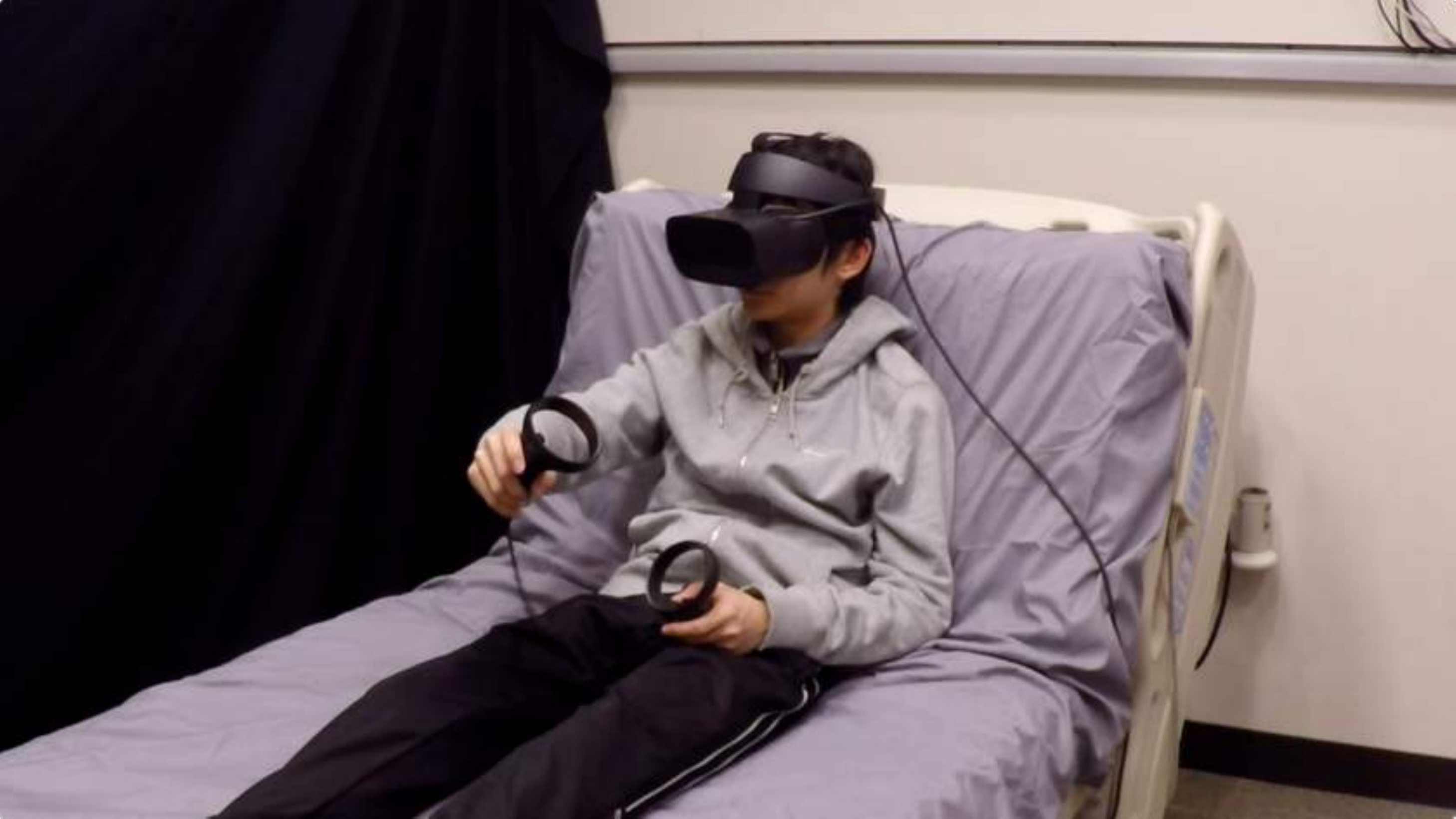}
\includegraphics[width=0.24\textwidth, trim={3cm 3cm 3cm 2cm}, clip]{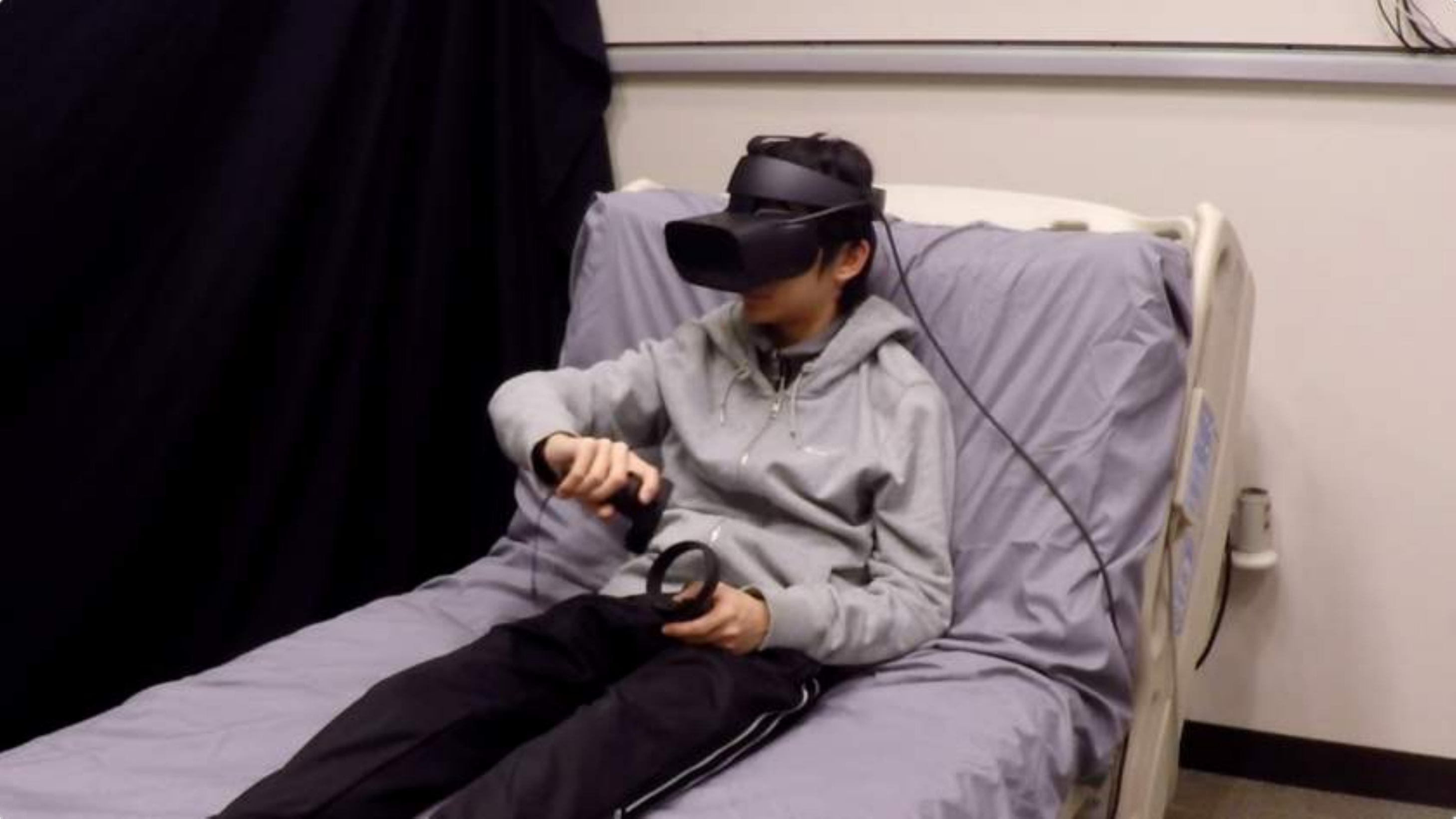}
\includegraphics[width=0.24\textwidth, trim={3cm 3cm 3cm 2cm}, clip]{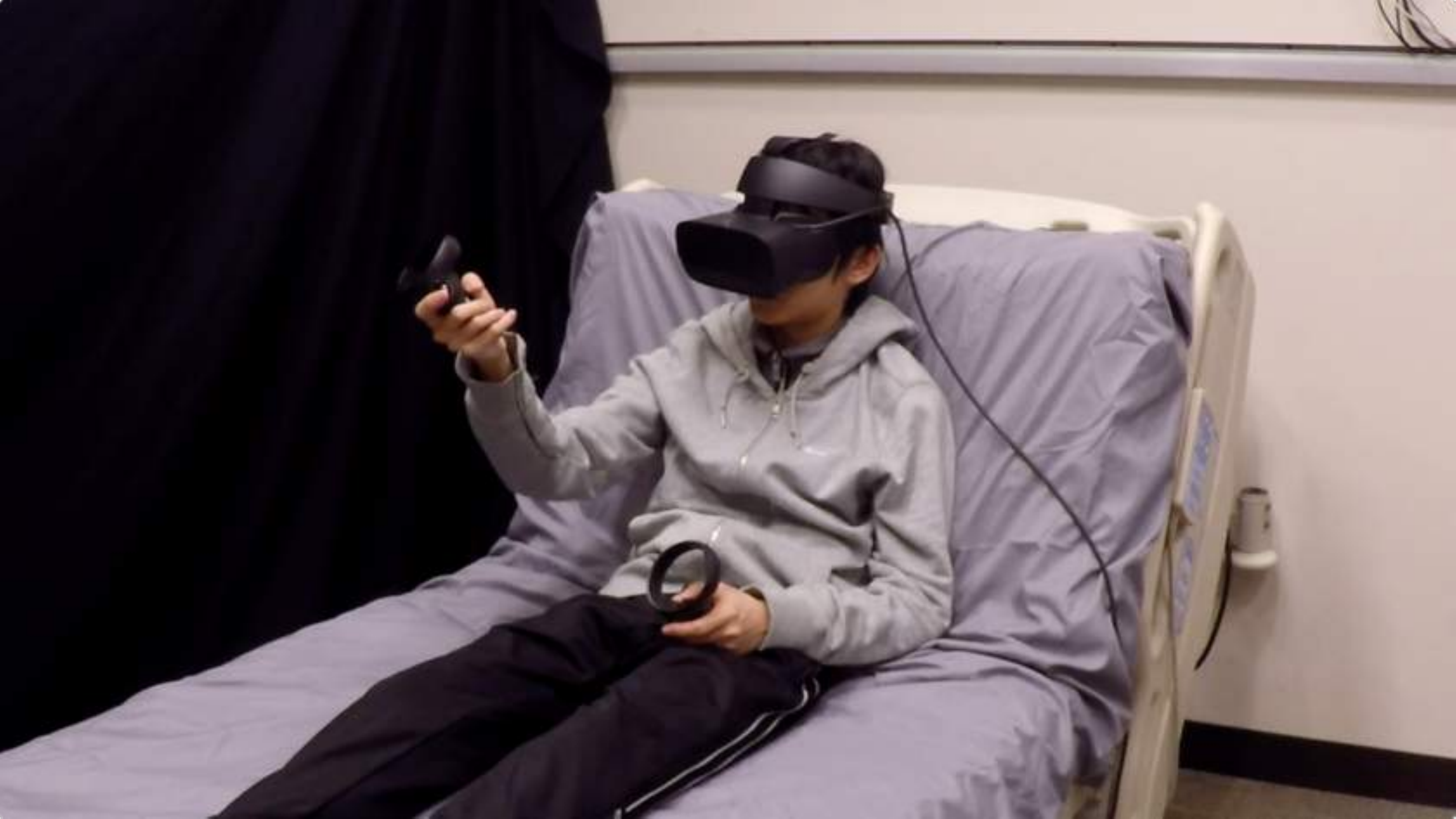}\\
\ \includegraphics[width=0.24\textwidth, trim={3cm 4cm 3cm 1cm}, clip]{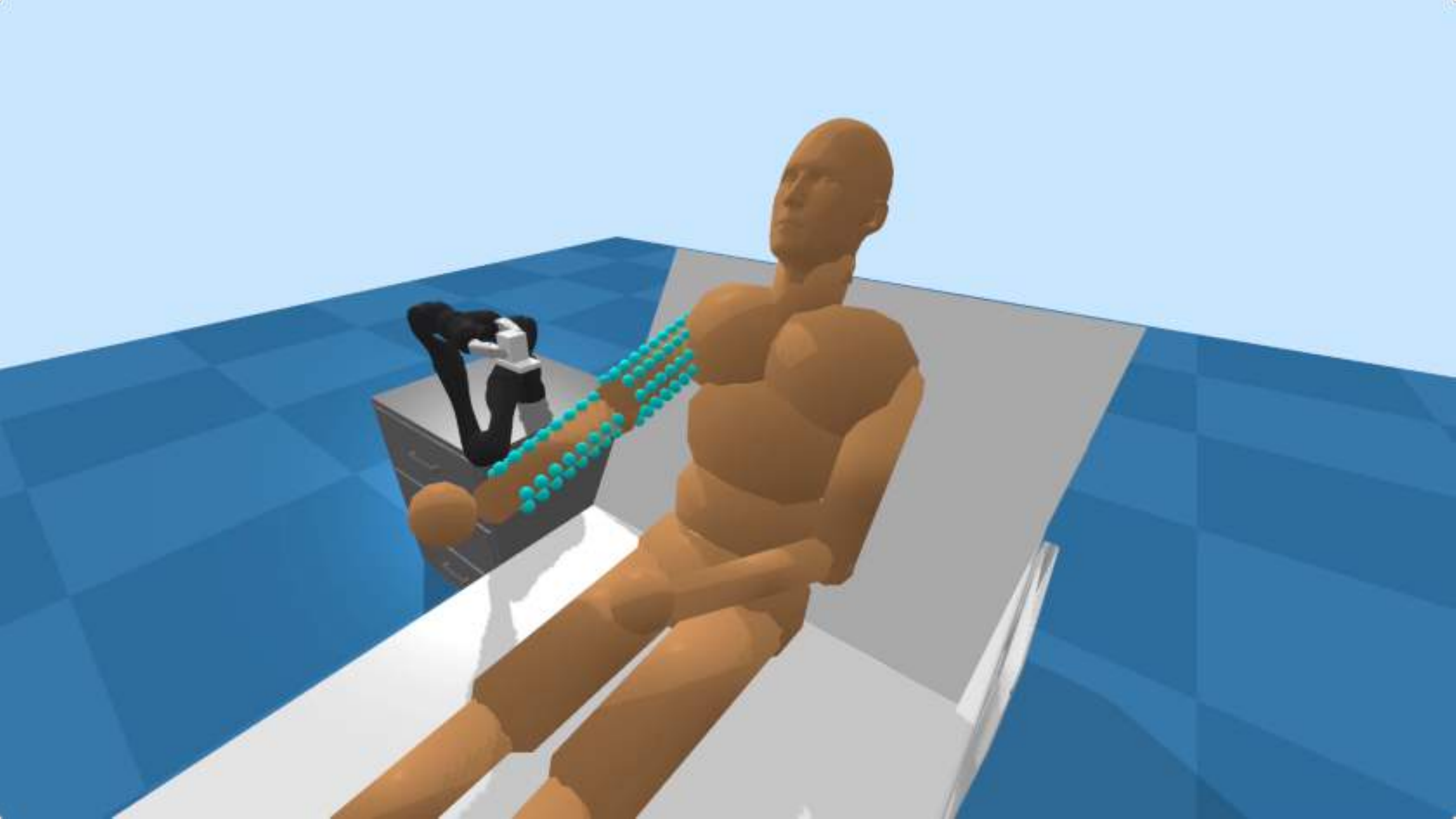}
\includegraphics[width=0.24\textwidth, trim={3cm 4cm 3cm 1cm}, clip]{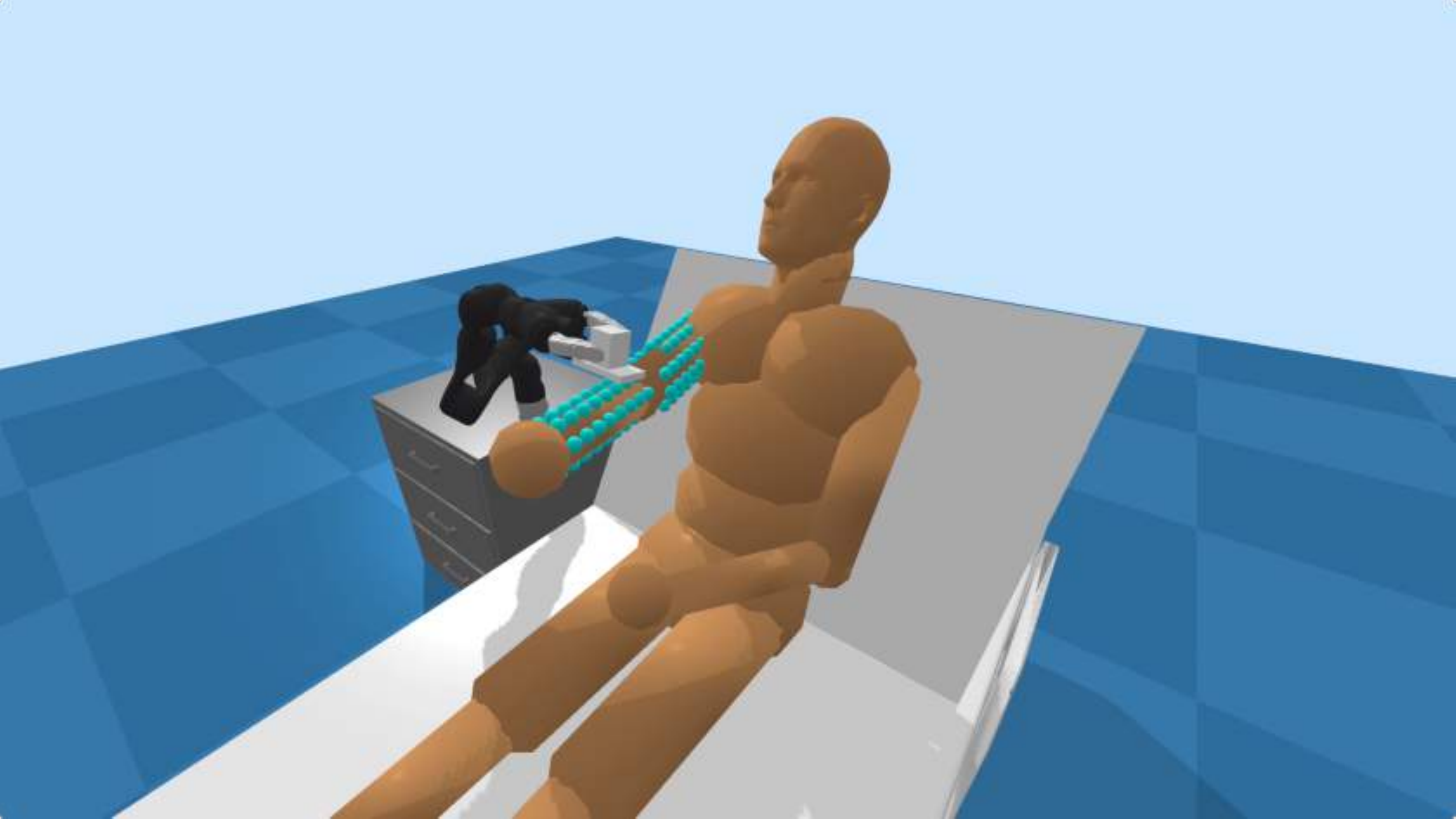}
\includegraphics[width=0.24\textwidth, trim={3cm 4cm 3cm 1cm}, clip]{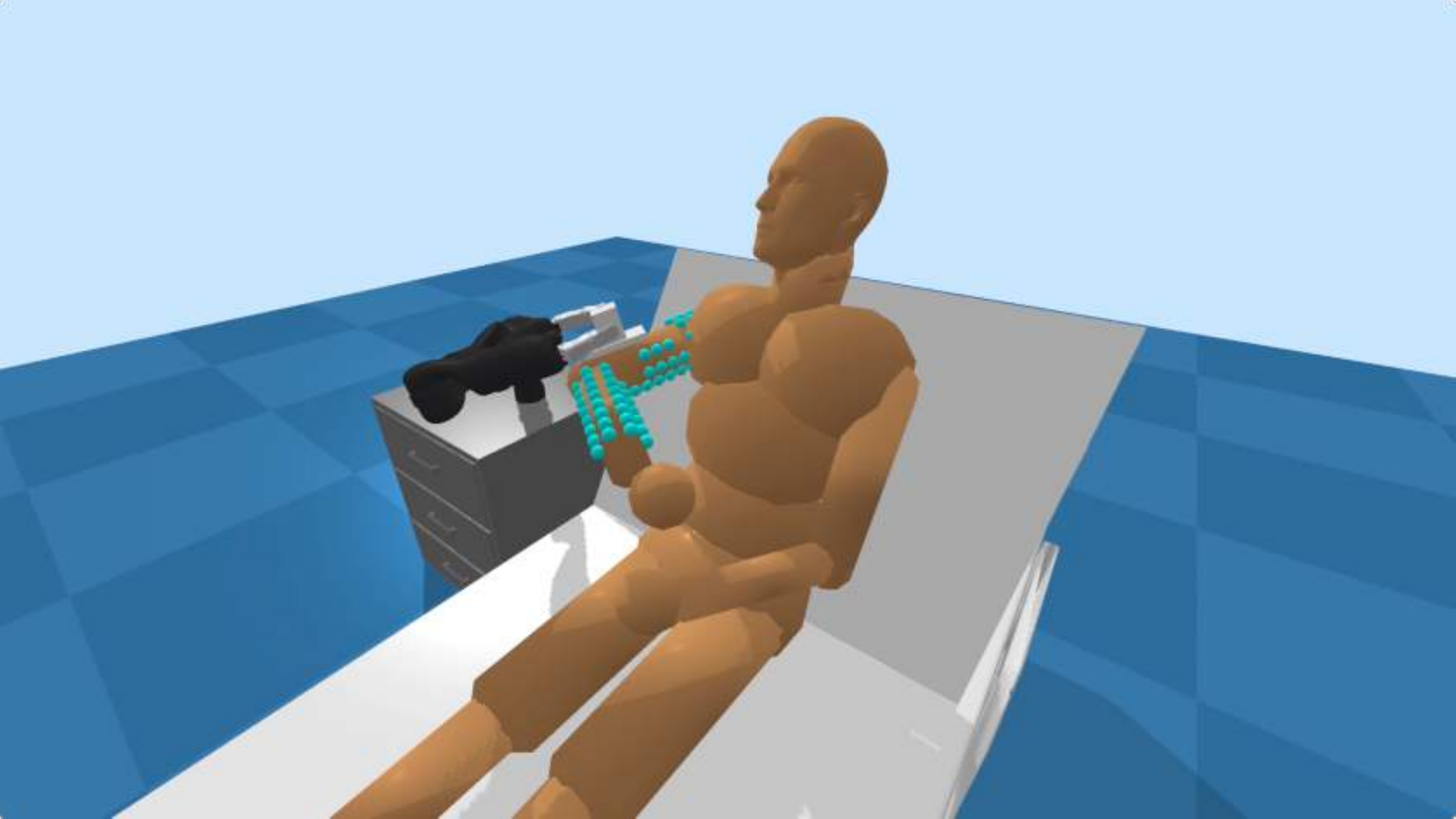}
\includegraphics[width=0.24\textwidth, trim={3cm 4cm 3cm 1cm}, clip]{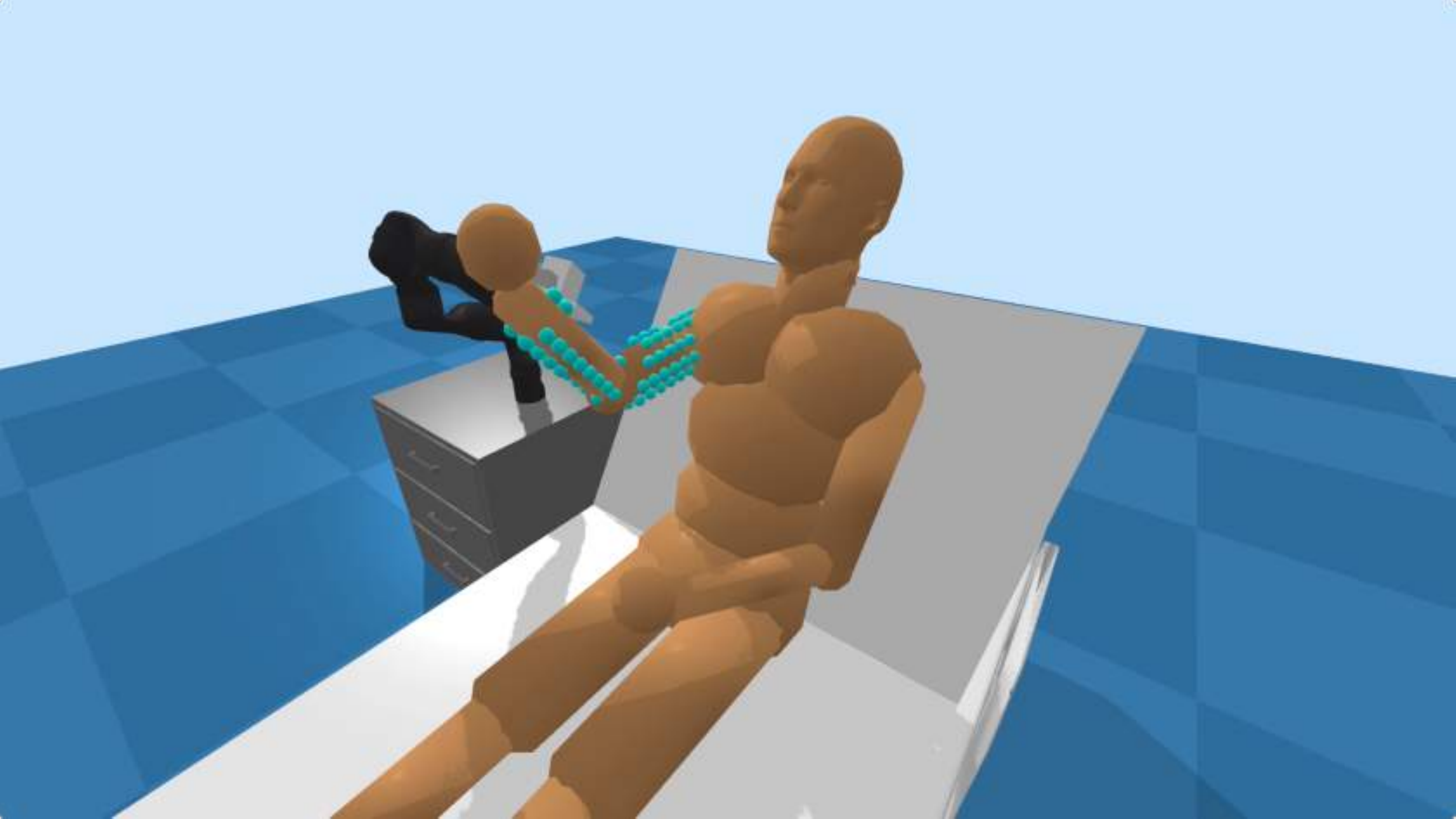}
\vspace{-0.2cm}
\caption{\label{fig:bed_bathing}Image sequence of a table-mounted Jaco providing bed bathing assistance using the Revised control policy.}
\vspace{-0.2cm}
\end{figure*}

\begin{figure*}
\centering
\includegraphics[width=0.24\textwidth, trim={8cm 6.5cm 7cm 2.5cm}, clip]{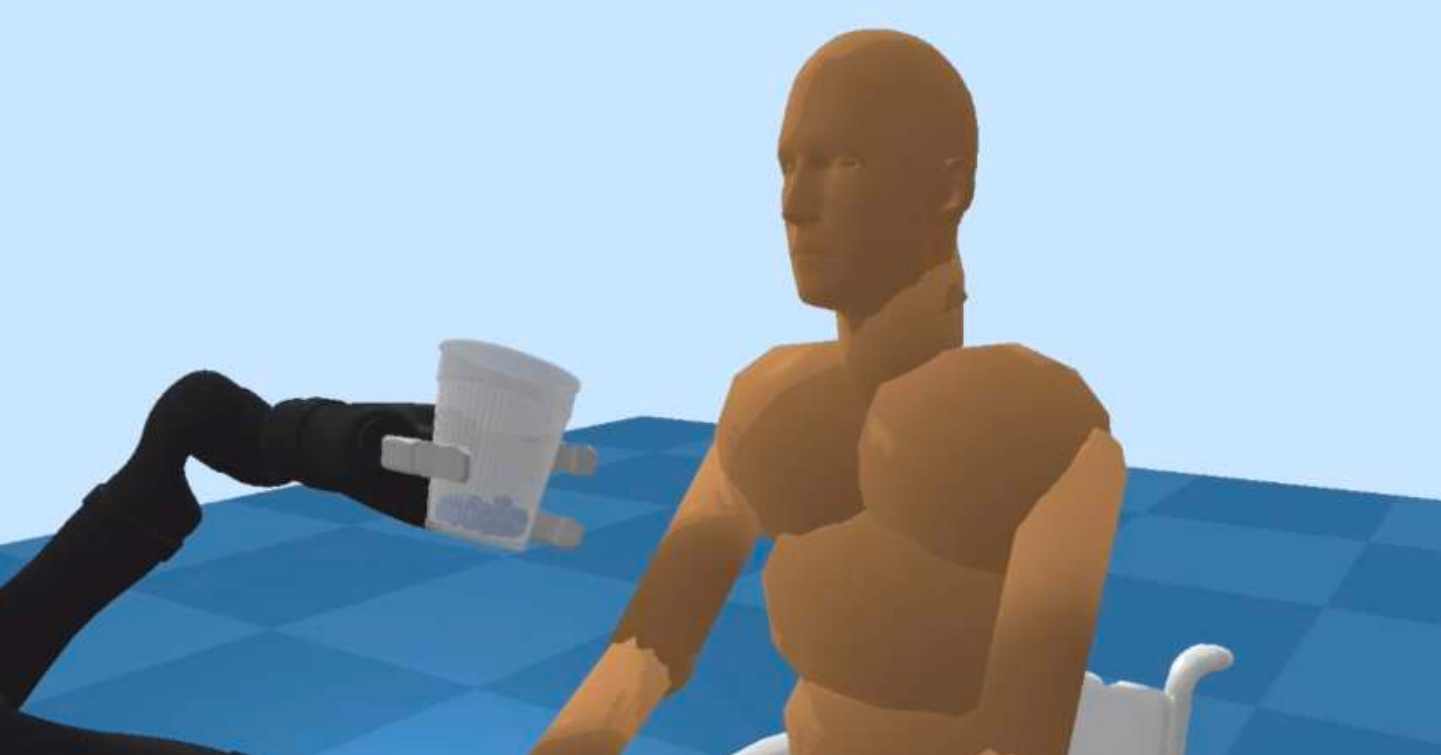}
\includegraphics[width=0.24\textwidth, trim={8cm 6.5cm 7cm 2.5cm}, clip]{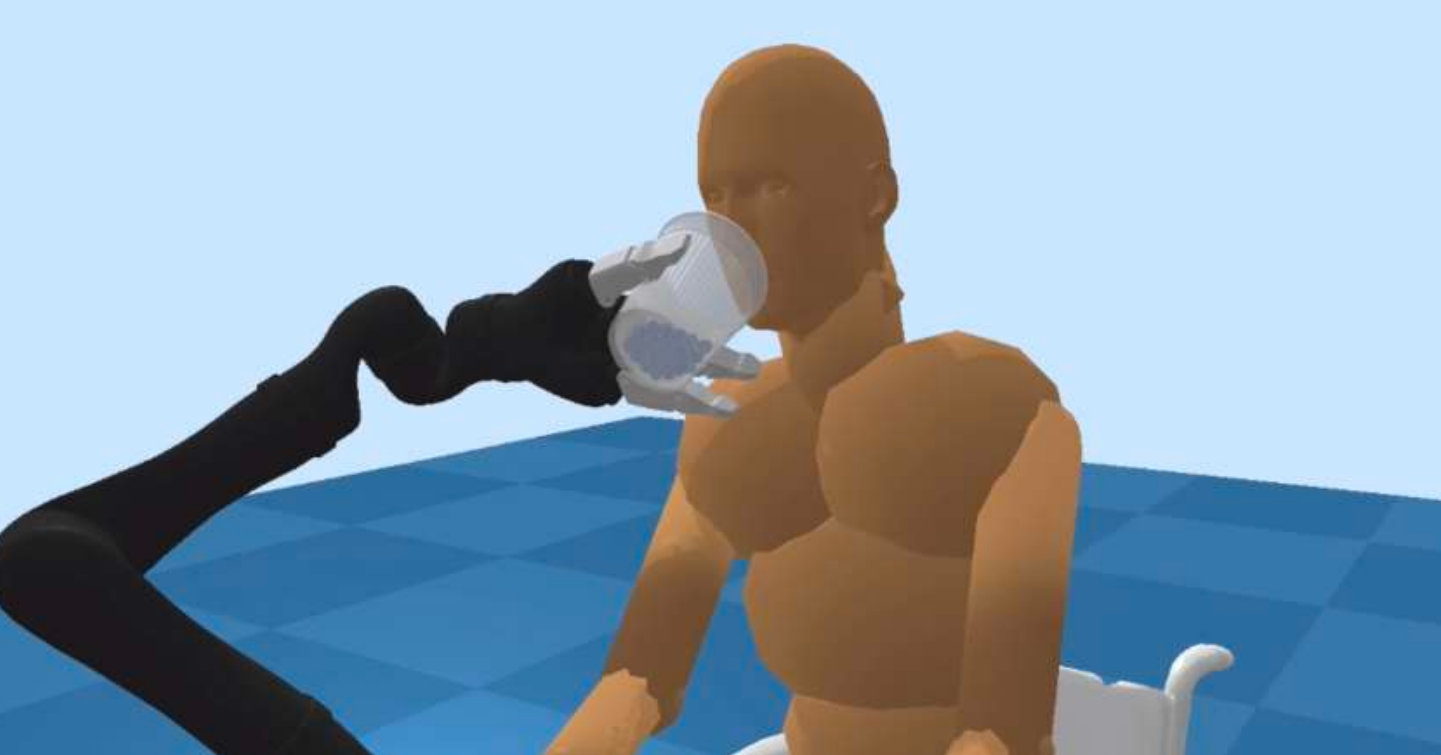}
\includegraphics[width=0.24\textwidth, trim={8cm 6.5cm 7cm 2.5cm}, clip]{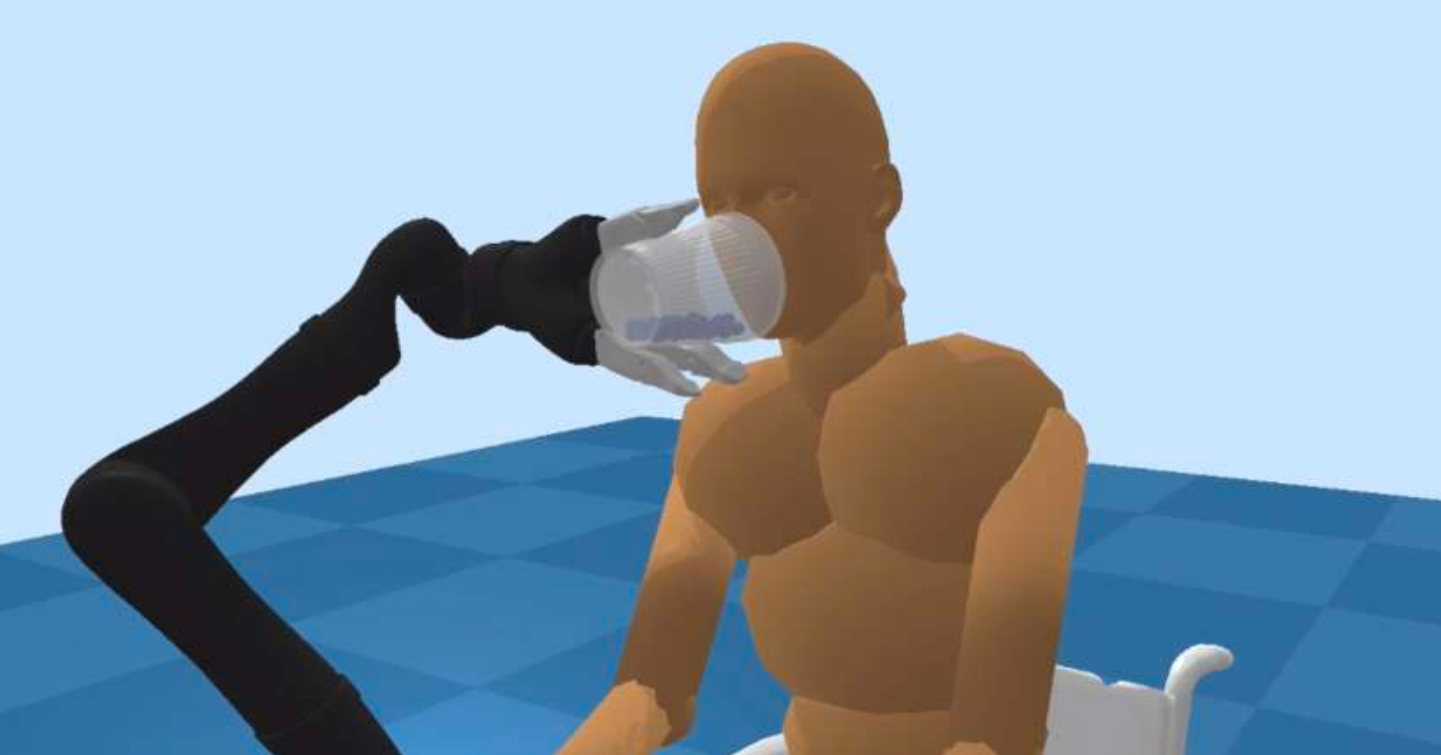}
\includegraphics[width=0.24\textwidth, trim={8cm 6.5cm 7cm 2.5cm}, clip]{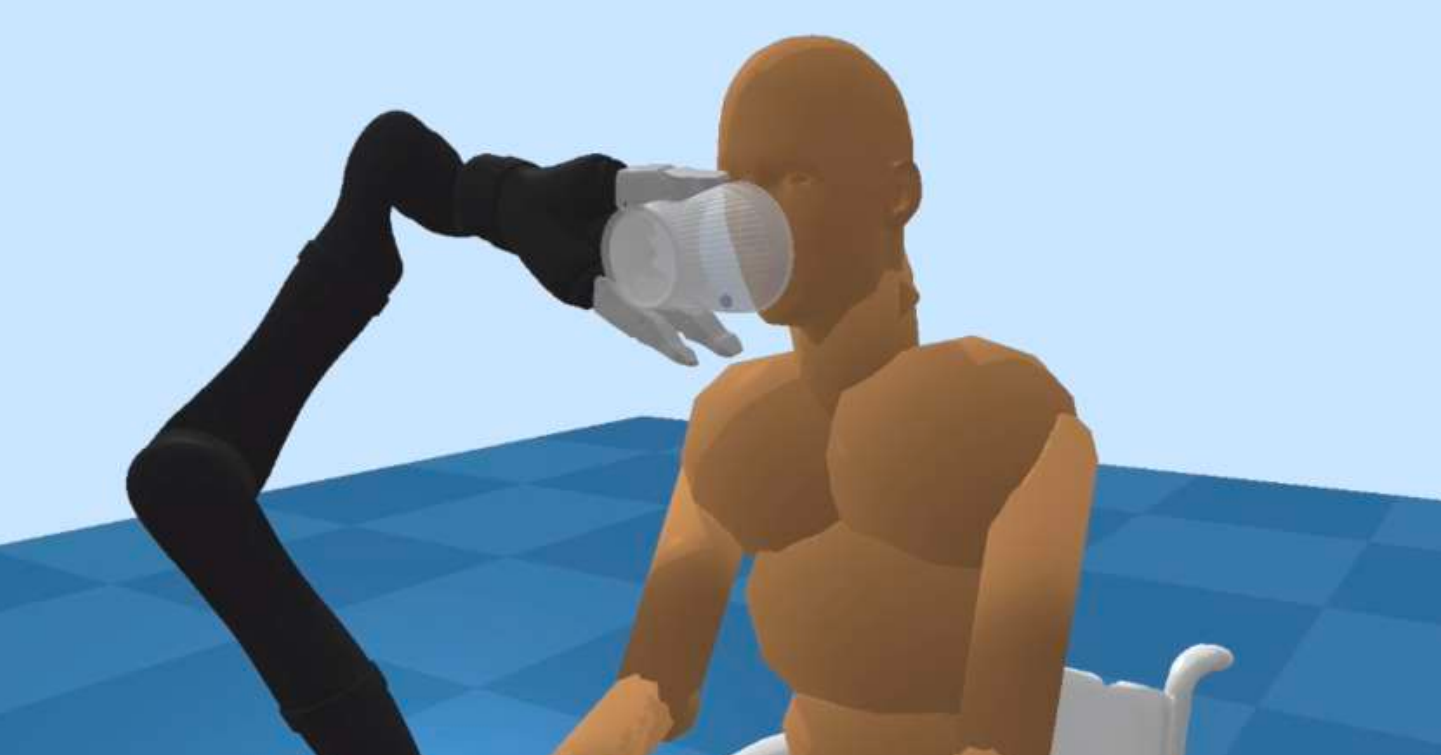}\\
\ \includegraphics[width=0.24\textwidth, trim={8cm 6.5cm 7cm 2.5cm}, clip]{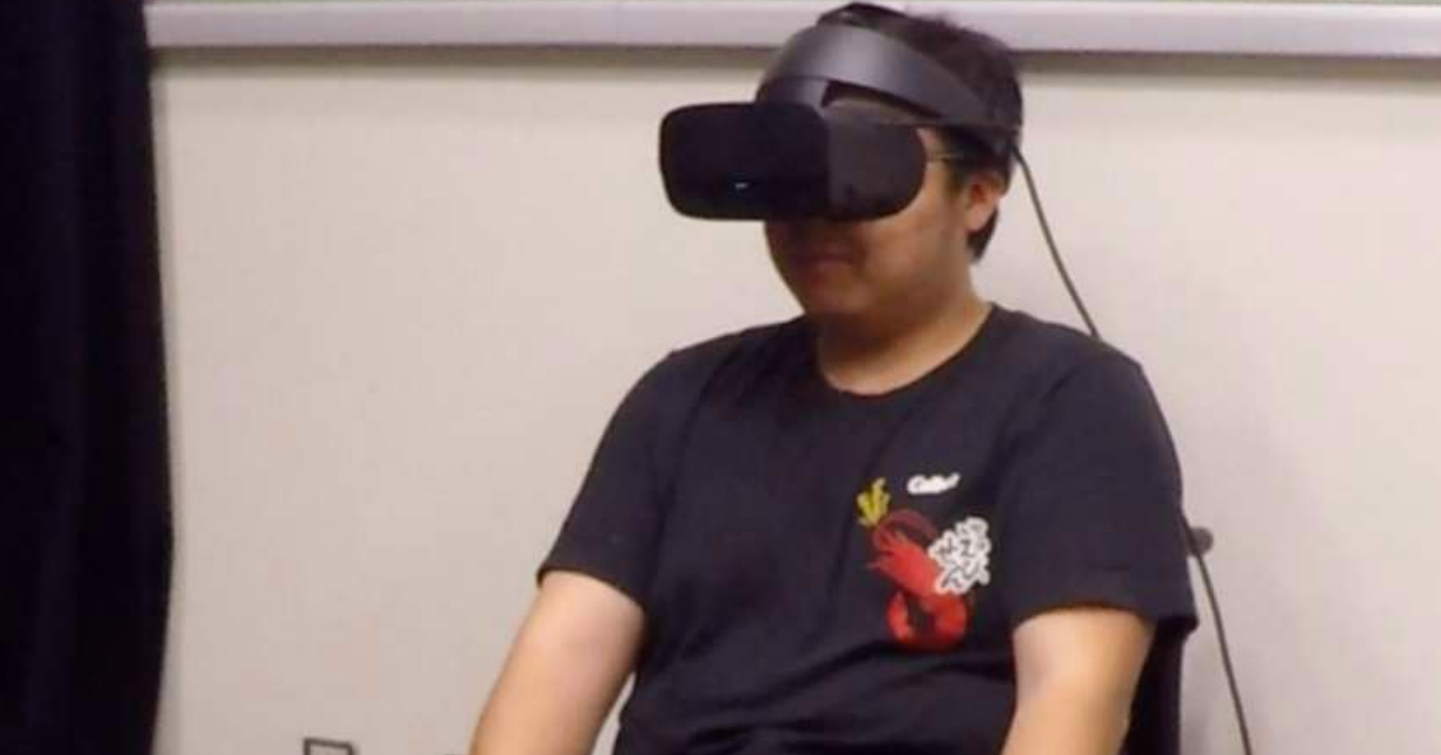}
\includegraphics[width=0.24\textwidth, trim={8cm 6.5cm 7cm 2.5cm}, clip]{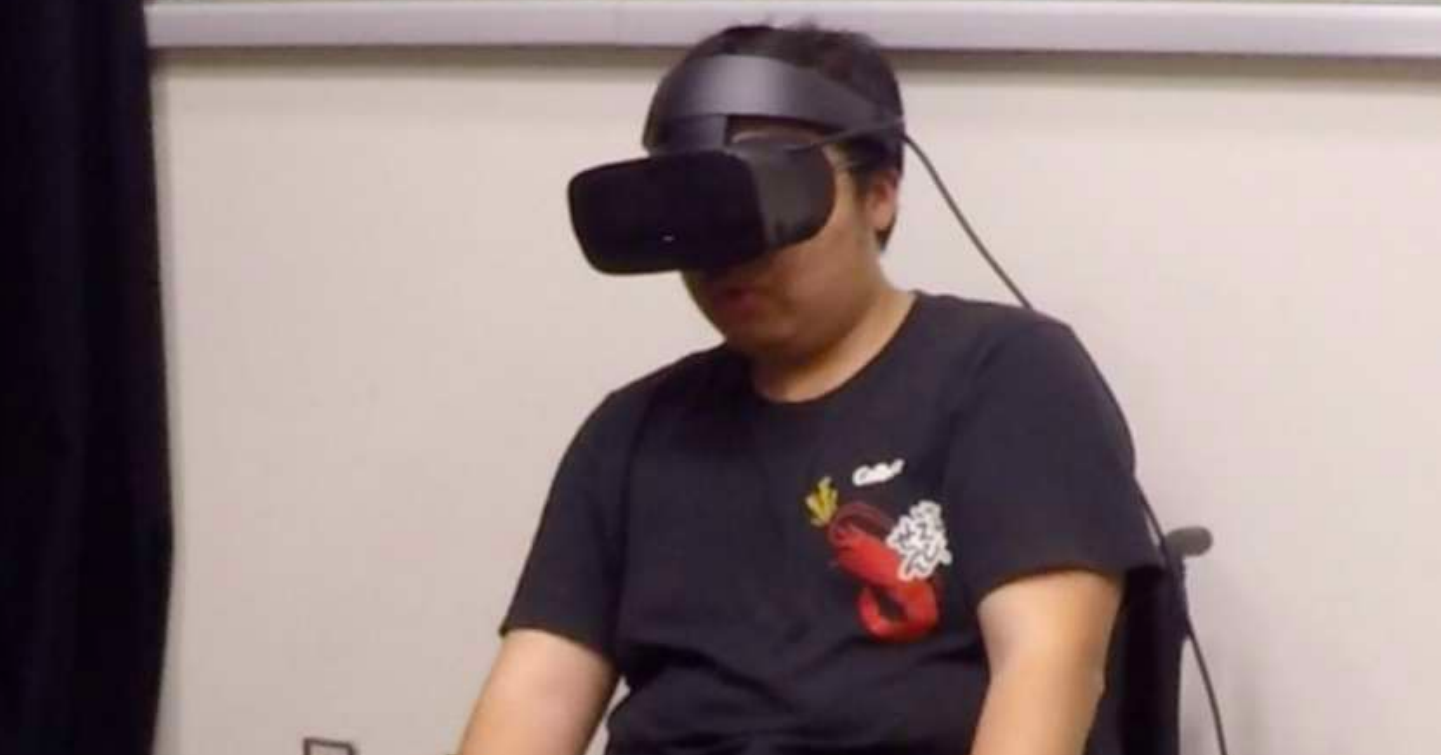}
\includegraphics[width=0.24\textwidth, trim={8cm 6.5cm 7cm 2.5cm}, clip]{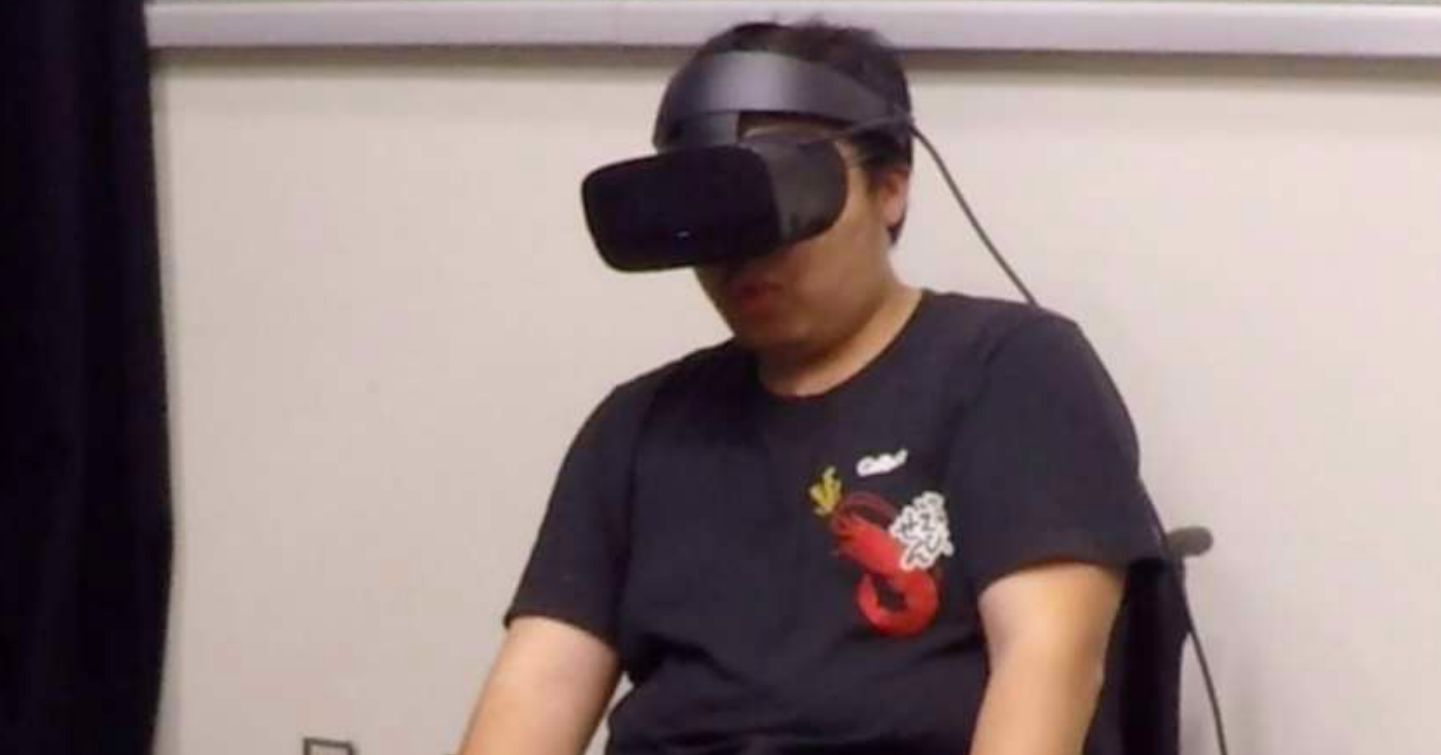}
\includegraphics[width=0.24\textwidth, trim={8cm 6.5cm 7cm 2.5cm}, clip]{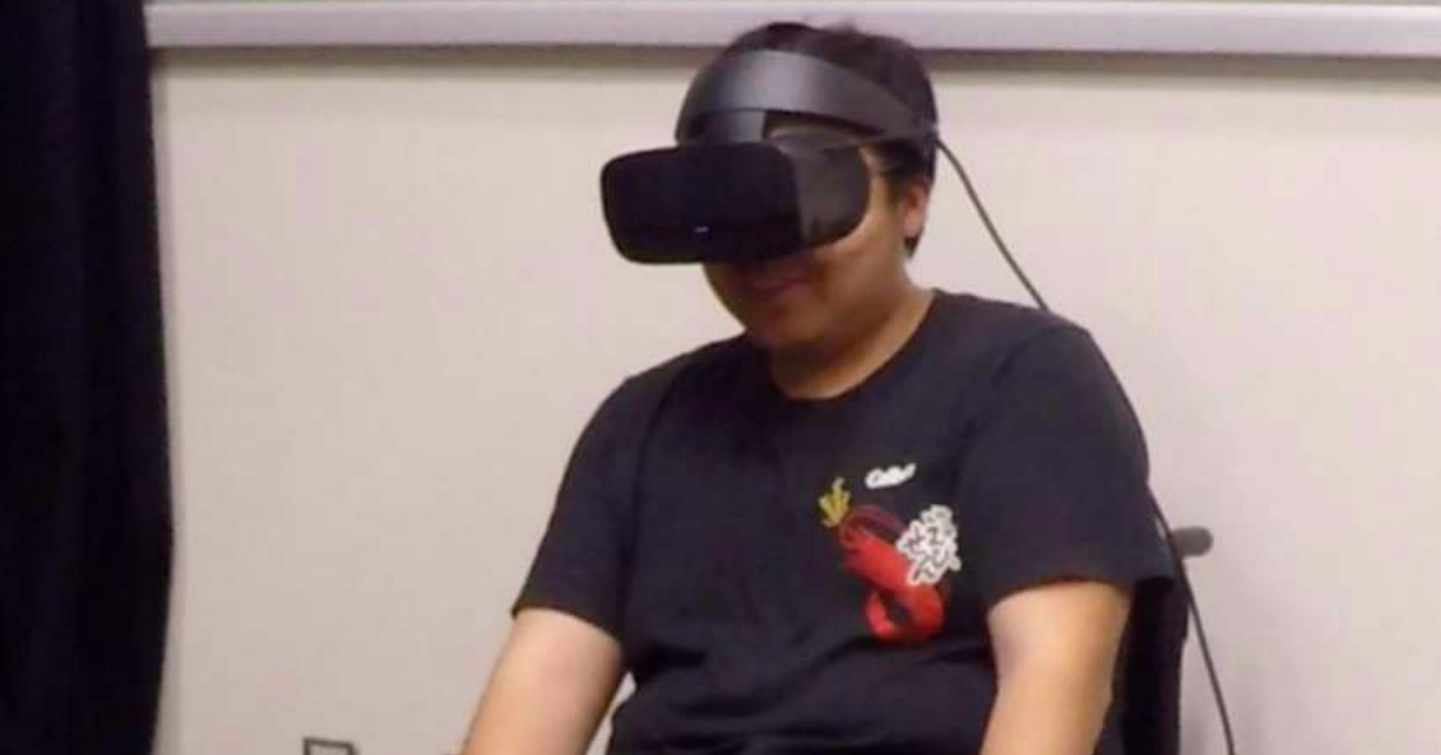}
\vspace{-0.2cm}
\caption{\label{fig:drinking}Image sequence of drinking assistance with the Jaco and the Revised control policy.}
\vspace{-0.5cm}
\end{figure*}

\begin{figure}
\centering
\includegraphics[width=0.45\textwidth, trim={0cm 0cm 0cm 2.5cm}, clip]{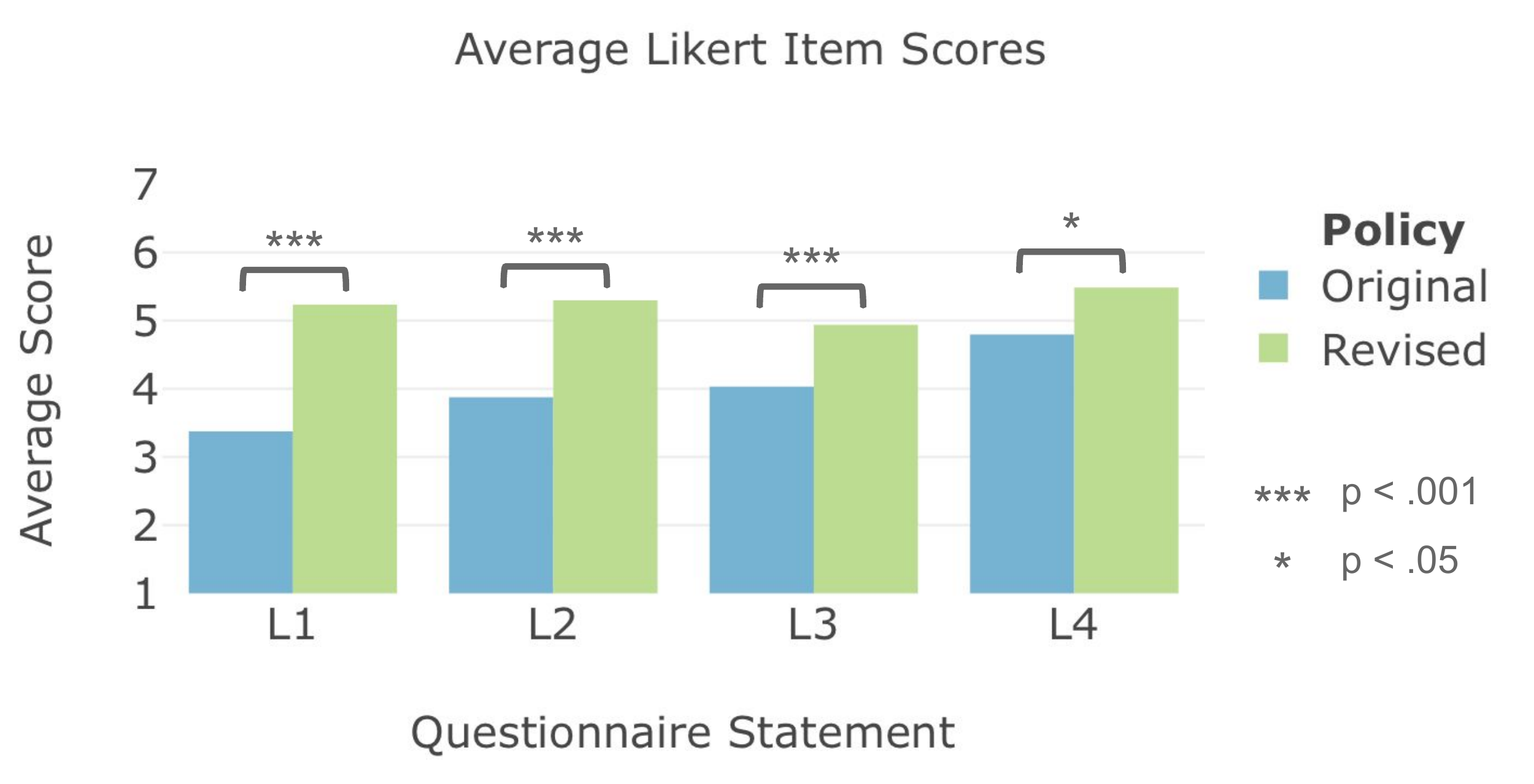}
\vspace{-0.2cm}
\caption{\label{fig:likert}Comparison of 7-point scale questionnaire responses for both Original and Revised policies. Responses are averaged over all participants, tasks, and robots. p-values are computed with a Wilcoxon signed-rank test.}
\vspace{-0.5cm}
\end{figure}

Table~\ref{table:original_improved_compare} compares the average reward and task success between the Original and Revised control policies during our VR study. For a given task and robot, we average the rewards and task success rates over trials from all eight participants.

From the table, we observe that the Revised policies for both robots outperformed the Original policies when providing itch scratching and bed bathing assistance.
Fig.~\ref{fig:bed_bathing} demonstrates the Jaco robot using a Revised policy to provide bed bathing assistance to a participant in virtual reality. 
While the Original policies for the Jaco robot failed to reliably assist human participants with drinking, we observe that the Revised policies for the Jaco exhibited a greater level of success.
Fig.~\ref{fig:drinking} depicts an image sequence of the Jaco robot using a Revised control policy to provide drinking assistance.
Based on these success metrics and qualitative image sequences, we find that the Revised polices exhibited reasonable performance for all of the assistive tasks.
However, we note that these revised controllers are not perfect and still exhibit some errors. This is especially true for hard to reach locations, such as scratching an itch underneath a person's arm. These control policies also assume that the person receiving assistance acts collaboratively with the robot. Additional research is required to explore how robot controllers learned in simulation can act appropriately when exposed to uncooperative or adversarial human motion.

For all four statements (L1, L2, L3, and L4), participants' average responses were positive or neutral (see Fig.~\ref{fig:likert}). Most notably, participants tended to perceive the robots as being successful at assistance, as evidenced by responses to L1 being significantly above neutral (4) with $p<0.001$ using a Wilcoxon signed-rank test.

Fig.~\ref{fig:likert} displays a comparison of questionnaire responses between the Original and Revised policies when responses are averaged over all participants, tasks, and robots for each questionnaire statement. On average, participants provide higher responses for trials in which a robot used a Revised policy to provide assistance. To determine a statistical difference between the Original and Revised policies, we apply a Wilcoxon signed-rank test, a non-parametric test, between participant responses from trials using the Original control policies and responses when using the Revised policies. The computed p-values are depicted in Fig.~\ref{fig:likert}, where we observed a statistically significant difference at the $p<.001$ level for the first three questions relating to task success, comfort, and safety. 
These results indicate that participants perceived a noticeable improvement in a robot's performance and safety when using Revised control policies that were trained based on biomechanics feedback from virtual reality.

Overall, we found that that our Revised policies significantly outperformed the Original policies, which indicates that VR can be used to efficiently improve simulation-trained controllers for physical human-robot interaction.

\subsection{Posthoc Biomechanical Analysis}

Our iterative research and development using AVR Gym, discussed in Section~\ref{sec:improved_policies}, resulted in Revised policies trained with new distributions of human torso heights and waist orientations. Here, we provide a posthoc statistical analysis of these biomechanical parameters for our simulated humans and our real participants in virtual reality. 

During our human study, we aligned the human model in virtual reality with a real participant's pose, as described in Section~\ref{sec:vr}. This alignment enabled us to record the full state, height, and pose of a participant at each time step during the virtual reality trials. 

To determine whether a statistical difference exists between human torso heights, we applied a Wilcoxon signed-rank test between human torso heights in the simulation environments used to train the Original policies and torso heights of human participants in virtual reality. This test returned a p-value of $p<0.001$, indicating that the original simulation environments did not properly model the distribution of real human heights observed during virtual reality. 

We then applied the same test to a set of torso heights from the updated simulation environments used to train the Revised policies. When compared to torso heights of human participants, the test returns a p-value of $p=0.29$, which indicates that there is a less significant difference between the distributions of human heights used to train the Revised control policies and human heights observed in virtual reality.

In order to evaluate differences in waist orientations between simulation and virtual reality, we consider the 3D position of a person's head, averaged over time.
Since the human model in virtual reality is aligned with a real participant's pose, we track the 3D head position of a participant by querying the 3D head position of the virtual human model.
In order to separate biomechanical differences in torso height from waist orientations, we modified the original simulation environments to include random variation among human torso heights, but no variation among waist orientations.
Using a Wilcoxon signed-rank test, we found a statistically significant difference in head position (averaged over all time steps for a trial) between data from these new simulation environments and the head positions of participants in virtual reality, with a p-value of $p<0.01$.
However, by performing this same test between head positions from the revised simulation environments (Section~\ref{sec:improved_policies}) and from VR, we found a p-value of $p=0.26$, indicating that the improved simulation environments used to train our Revised policies better matched the biomechanics of our participants in VR.

\section{Conclusion}

We presented Assistive VR Gym (AVR Gym), which enables real people to interact with virtual assistive robots. We also provided evidence that AVR Gym can help researchers improve the performance of simulation-trained assistive robots. Overall, our results suggest that VR can be used to help bridge the gap between physics simulations and the real world by enabling real people to interact with virtual robots. 

\section*{Acknowledgment}

\small
\textit{We thank Pieter Abbeel, Anca Dragan, and Deepak Pathak for feedback about this work. This work was supported by NSF award IIS-1514258 and AWS Cloud Credits for Research. Dr. Kemp owns equity in and works for Hello Robot, a company commercializing robotic assistance technologies.}

\bibliographystyle{IEEEtran}
\bibliography{bibliography}

\end{document}